%% file: main.tex
\documentclass{article}

\usepackage{arxiv}

\usepackage[utf8]{inputenc} 
\usepackage[T1]{fontenc} 
\usepackage{hyperref} 
\usepackage{url} 
\usepackage{booktabs} 
\usepackage{amsfonts} 
\usepackage{nicefrac} 
\usepackage{microtype} 
\usepackage{multirow}
\usepackage{amsmath}
\usepackage{amsfonts}
\usepackage{cleveref} 
\usepackage{lipsum} 
\usepackage{graphicx}
\usepackage{natbib}
\usepackage{doi}
\usepackage{xcolor}
\usepackage{dirtree}

\newlength{\savedwidth}
\newcommand{\whline}[1]{\noalign{\global\savedwidth\arrayrulewidth \global\arrayrulewidth #1}%
\hline \noalign{\global\arrayrulewidth\savedwidth}}

\title{EasyChauffeur: A Baseline Advancing Simplicity and Efficiency on Waymax}

\date{}

\author{Lingyu Xiao$^{1,2}$\thanks{Equal contribution. $^\dag$ Work done during an internship at Baidu. 
$^\ddag$ Corresponding author.}$^{\dag}$
, Jiang-Jiang Liu$^{2*}$, Xiaoqing Ye$^{2}$, Wankou Yang$^{1\ddag}$
, Jingdong Wang$^{2}$ \\ \\ $^{1}$Southeast University \quad $^{2}$Baidu }

\renewcommand{\shorttitle}{EasyChauffeur: A Baseline Advancing Simplicity and Efficiency
on Waymax}

\hypersetup{
	pdftitle={EasyChauffeur: A Baseline Advancing Simplicity and Efficiency on Waymax},
	pdfauthor={Lingyu Xiao, Jiang-Jiang Liu, Xiaoqing Ye, Wankou Yang, Jingdong Wang},
}

\begin{document}
	\maketitle

	\begin{abstract}
		Recent advancements in deep-learning-based driving planners have primarily focused
		on elaborate network engineering, yielding limited improvements. This paper diverges
		from conventional approaches by exploring three fundamental yet underinvestigated
		aspects: training policy, data efficiency, and evaluation robustness. We introduce
		EasyChauffeur, a reproducible and effective planner for both imitation
		learning (IL) and reinforcement learning (RL) on Waymax, a GPU-accelerated simulator.
		Notably, our findings indicate that the incorporation of on-policy RL significantly
		boosts performance and data efficiency. To further enhance this efficiency, we
		propose SNE-Sampling, a novel method that selectively samples data from the
		encoder's latent space, substantially improving EasyChauffeur's performance with
		RL. Additionally, we identify a deficiency in current evaluation methods,
		which fail to accurately assess the robustness of different planners due to
		significant performance drops from minor changes in the ego vehicle's
		initial state. In response, we propose Ego-Shifting, a new evaluation setting
		for assessing planners' robustness. Our findings advocate for a shift from a
		primary focus on network architectures to adopting a holistic approach encompassing
		training strategies, data efficiency, and robust evaluation methods. 
	\end{abstract}


	\section{Introduction}

	\begin{figure*}[tp]
		\includegraphics[width=\linewidth]{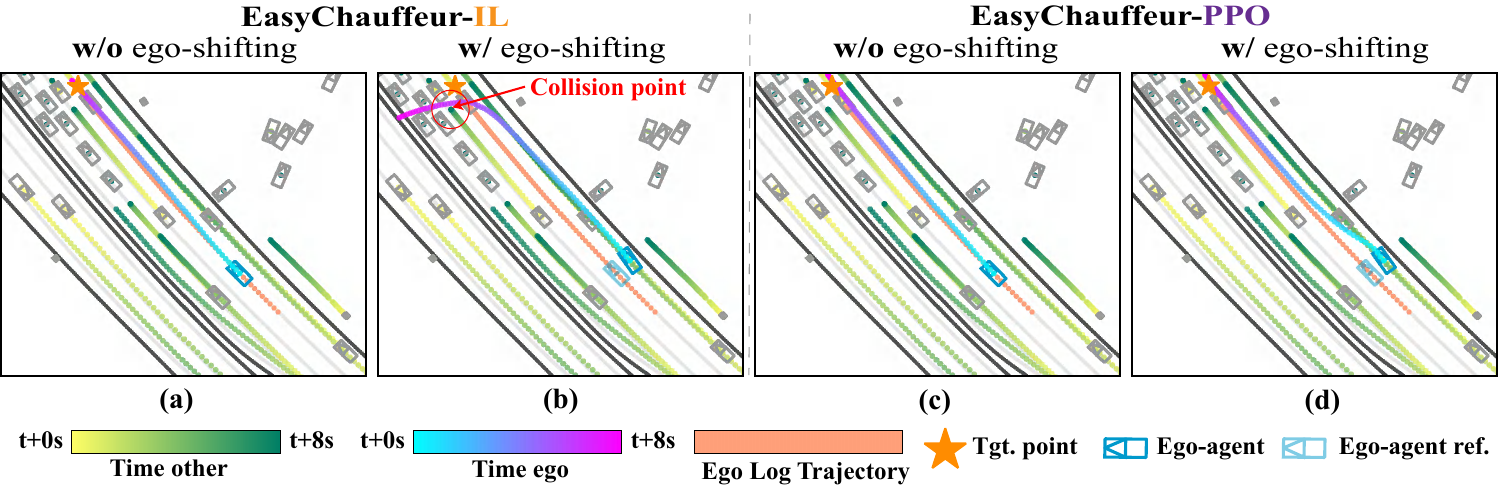}
		\caption{Comparison of the robustness of EasyChauffeur-IL and EasyChauffeur-PPO
		to the initial state under \textit{close-loop} evaluation. `Tgt. point' stands
		for the target point. `ego agent ref.' refers to the initial state of the ego agent
		used as a reference when evaluated under Ego-Shifting. The current setting
		for evaluation may not fully assess the planners' robustness, as shown in (a)
		and (c) where both models arrived successfully. However, when evaluated
		under Ego-Shifting, EasyChauffeur-IL (b) experiences a collision, while
		EasyChauffeur-PPO (d) demonstrates strong robustness. More visualisation results
		can be found on supplementary materials.}
		\label{fig.teaser}
	\end{figure*}

	The design of autonomous driving systems typically comprises three main
	components: prediction, perception, and planning. Recent advancements in deep
	learning techniques have greatly contributed to the progress of prediction
	tasks, revolutionising the onboard perception system. The planning module, which
	serves as the decision-making stage of the autonomous driving system, directly
	impacts overall performance and is rarely investigated in depth. One approach is
	to design the planning module along with the other two components in an end-to-end
	manner, meaning all modules are optimised jointly. However, this highly
	coupled design results in low explainability. Another approach is to design the
	planning module independently, known as the mid-to-end manner, where the perception
	module provides structured data to the planning module for generating control signals.
	In this paper, we focus on the latter approach.

	In the realm of planning for autonomous driving, metrics are computed in both
	\textit{close-loop} and \textit{open-loop} settings. During \textit{close-loop}
	evaluation, the ego vehicle’s states are computed via real-time interaction in
	the simulator, while in \textit{open-loop}, they are merely replayed through
	predetermined logs. This characteristic has motivated previous methods to embrace
	Imitation Learning (IL), treating planning as a supervised regression task
	since \textit{open-loop} emphasises the difference between prediction and replayed
	logs.

	As the field of driving has matured, to improve performance in \textit{close-loop},
	most works have focused on network engineering by incorporating some degree of
	environmental interaction. Examples include combining rule-based planners with
	learning-based ones~\cite{pdm}, introducing Monte-Carlo Tree Search (MCTS) in
	the decision-making phase~\cite{chekroun2023mbappe}, or formulating the problem
	with hierarchical game theory~\cite{gameformer}. It is important to note that
	the accumulation of errors can significantly amplify over time steps during
	\textit{close-loop} evaluation, necessitating the ability to prevent safety
	hazards. However, given that the collected demonstrations for IL primarily
	consist of flawless trajectories without any explicit instruction on avoiding
	accidents, the potential for substantial improvements through intricate design
	is limited. On the other hand, even though Reinforcement Learning (RL) is inherently
	characterized by adequate interaction, none of the methods under this category,
	such as those referenced in~\cite{liu2022improved,huang2022efficient,lu2023imitation},
	demonstrate strong scalability, effectiveness, and high reproducibility on a
	standardized large-scale benchmark like nuPlan~\cite{caesar2021nuplan} due to
	time-consuming rollouts on CPUs. Although the recently released GPU-accelerated
	simulator Waymax~\cite{waymax}, initialized with over 250 hours of real
	driving data from the Waymo Open Motion Dataset (WOMD)~\cite{Ettinger_2021_ICCV},
	makes cost-effective large-scale training and evaluation possible, there is
	neither open-source training code available for the community to use, for IL or
	RL, nor a released model to compare with.

	In this paper, to address the aforementioned problems, we conduct our study
	from three fundamental yet previously overlooked dimensions: training policy,
	data efficiency, and evaluation robustness. We begin by introducing a simple
	yet effective baseline planner for the community, applicable to both IL and RL.
	‘Simple’ stands for straightforward design with all necessary components, while
	‘effective’ denotes performance that is comparable to or better than the reported
	metrics in~\cite{waymax}. Surprisingly, we find that introducing an on-policy RL
	strategy, namely Proximal Policy Optimisation (PPO)~\cite{ppo}, into planner training
	can greatly improve performance with only approximately 0.6\% of the data.

	Furthermore, with regard to data considerations, an ablation study on the
	volume of training data for IL reveals a phenomenon characterized by premature
	saturation with limited performance potential, suggesting that the data is redundant
	and easy to fit. Intuitively, we propose a sampling method called SNE-Sampling
	(Stochastic Neighbor Embedding-based Sampling), which operates on the latent
	space to select representative data. RL trained with data from SNE-Sampling has
	proven to have a significant performance gain compared to the vanilla approach.

	Lastly, during \textit{close-loop} evaluation, we find that performance in
	the simulator and the real-world is misaligned: the simulator lacks the imperfect
	initial localisation that is widely present in reality, for example, overtaking,
	pulling over, or localisation errors. The absence of such demonstrations in
	evaluations allows nearly perfect imitation to achieve good results instead of
	focusing on whether the model is robust enough to respond appropriately, as demonstrated
	in Fig.~\ref{fig.teaser}. Therefore, we rethink this sim-to-real gap by
	proposing an Ego-Shifting setting that can be integrated into any simulator. Experiments
	demonstrate that IL methods have much worse generalizability than PPO under this
	setting.

	Most importantly, these interesting findings collectively highlight the
	potential shift from conventional network engineering approaches to the design
	of training strategies, the efficiency of data use, and the exploration of
	different evaluation directions.

	The main contributions of this paper are summarised as follows:
	\begin{itemize}
		\item \textbf{Framework-wise:} We provide the community with a simple yet
			effective planner for IL and RL on the WOMD-driven Waymax.

		\item \textbf{Data-wise:} Extensive experiments prove that RL can be
			scalable with limited training data ($\sim 0.6\%$), and we propose SNE-Sampling
			to select representative training data.

		\item \textbf{Evaluation-wise:} To address the sim-to-real gap in \textit{close-loop}
			evaluation that overlooks the assessment of models’ robustness, we propose
			an Ego-Shifting evaluation setting that can be applied to any simulator.
	\end{itemize}

	\section{Related Work}

	\subsection{Ego-Planning with Imitation Learning}

	\paragraph{End-to-End.}
	The end-to-end planner aims to produce future trajectories directly from raw sensor
	input. A naive approach involves directly mapping control signals via a CNN~\cite{codevilla2018end,codevilla2019exploring}.
	Later works have managed to fuse multimodal information by performing planning
	in BEV space~\cite{chitta2021neat} or by incorporating LiDAR and camera data~\cite{chitta2022transfuser},
	encoding the map with VectorNet~\cite{zhang2021tip}. Other works closely integrate
	planning and perception. LAV~\cite{chen2022learning} leverages additional
	information from other vehicles for better reasoning. STP3~\cite{hu2022st} perceives
	spatial-temporal features and plans a safe maneuver. VAD~\cite{jiang2023vad}
	injects every vectorized element into the transformer and supervises it with
	certain constraints. UniAD~\cite{hu2023planning} utilises a systematic model design,
	connecting intermediate task nodes through query vectors and jointly
	optimising them. However, as all modules are highly coupled, future maintenance
	becomes challenging.

	\paragraph{Mid-to-End.}
	These approaches focus on generating control signals from post-perception
	results. For instance, PlanT~\cite{plant} utilised abstracted object-level
	representations for easy integration with existing perception algorithms. Unlike
	end-to-end planners, some works in this category have proven effective in real-world
	applications. SafetyNet~\cite{vitelli2022safetynet} incorporates a rule-based fallback
	layer to improve performance. UrbanDriver~\cite{scheel2022urban} was trained
	on a differentiable data-driven simulator built on perception outputs and high-fidelity
	HD maps from real-world data. SafetyPathNet~\cite{pini2023safe} selects a planning
	trajectory that minimises a cost considering safety and predicted probabilities.
	With the release of the first real-world large-scale standardised benchmark, nuPlan~\cite{caesar2021nuplan},
	recent works have conducted extensive studies using this benchmark. PDM~\cite{pdm}
	combined a rule-based planner with a learning-based approach, hotplan~\cite{hu2023imitation}
	utilised heatmap representation to predict future multimodal states, GameFormer~\cite{gameformer}
	treated interaction predictions as a game theory problem, and MBAPPE~\cite{chekroun2023mbappe}
	introduced MCTS in the decision-making phase. Additionally,~\cite{cheng2023rethinking}
	argues that \textit{close-loop} and \textit{open-loop} system evaluations are
	misaligned and proposes essential components for \textit{close-loop} evaluation.
	However, all these IL-trained works may experience covariate shift and causal confusion.

	\subsection{Ego-Planning with Reinforcement Learning}

	Within the scope of end-to-end planners, CIRL~\cite{liang2018cirl}, built upon
	the Deep Deterministic Policy Gradient (DDPG) algorithm~\cite{lillicrap2015continuous},
	was the first RL model applied to CARLA~\cite{dosovitskiy2017carla}. Inspired
	by this,~\cite{kendall2019learning} successfully employed DDPG in a real vehicle.
	To address the insufficiency of gradients obtained via RL,~\cite{toromanoff2020end}
	successfully applied RL in CARLA when combined with IL. However, no reports have
	yet shown that RL outperforms IL in end-to-end planning~\cite{chen2023end}.
	The situation changes when privileged information is available. Roach~\cite{roach}
	was trained using off-policy PPO with privileged BEV semantic segmentation,
	and the resulting dataset was used to train an IL agent, achieving superior performance.~\cite{isele2018navigating}
	employed Deep Q-Network (DQN) with discretized BEV views to navigate a vehicle
	through intersections and occlusions.~\cite{wang2018reinforcement} modified the
	Q-function network structure to accommodate a continuous action space for performing
	lane changes. While improvements have been achieved by ~\cite{liu2022improved,huang2022efficient,lu2023imitation},
	and ~\cite{isele2018navigating,wang2018reinforcement}, they have focused on
	specific scenarios or settings rather than conducting extensive evaluations on
	a large-scale benchmark.
 
	\section{Approach}

	\begin{figure}[t]
		\centering
		\includegraphics[width=0.9\linewidth]{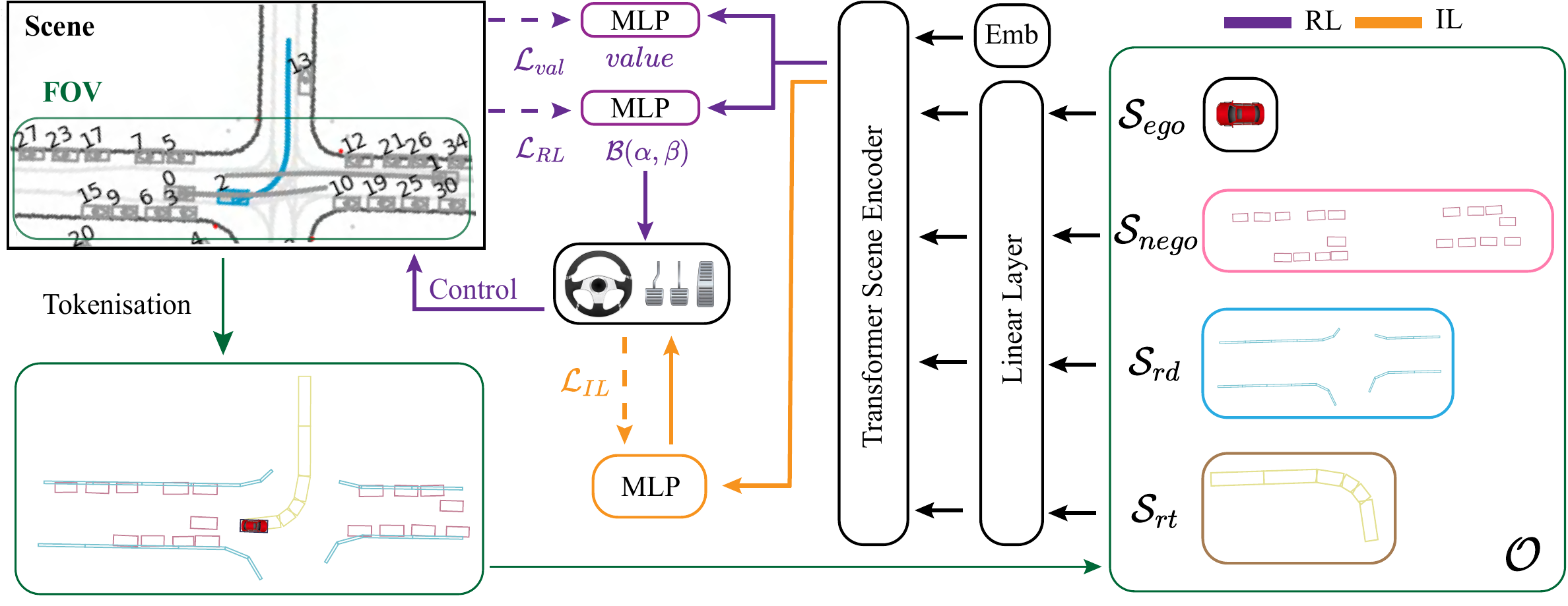}
		\caption{ Overall pipeline of EasyChauffeur. }
		\label{fig.pipeline}
	\end{figure}

	\subsection{EasyChauffeur}
	\paragraph{Network Design.}
	The overall network structure is illustrated in Fig.~\ref{fig.pipeline}. The
	design is straightforward and comprises three parts: Tokenisation, Transformer
	scene encoder, and MLP. The privileged information provided by the simulator (detection
	results when deployed in the real world) will be tokenised as a set of
	rectangles and fed into a transformer scene encoder separately by its own type.
	Since the input is one-dimensional and for simplicity, we utilise BERT~\cite{devlin2018bert}
	as the encoder. Here, a learnable embedding is introduced to fuse the
	privileged scene information via self-attention, thus serving as a latent space.
	After that, the fused feature is fed into an MLP that decodes the information into
	actual control signals $\mathcal{A}$, e.g., steering and acceleration when
	using the bicycle action space, or the desired target point for the next timestamp
	when using the waypoints action space.

	\paragraph{Input / Output Representation.}
	\label{sec.in/out_rep} The raw privileged scene information from Waymax is
	organised as follows: agents' trajectories and properties (where agents are the
	collection of dynamic objects within a scenario, e.g., vehicles, pedestrians,
	bicyclists), road map, and routing for the ego agent. As described in~\cite{waymax},
	the routing is the union of expert log trajectories and drivable futures.
	However, WOMD does not release the drivable futures; we only use expert log
	trajectories as the ego agent's routing. Next, we will introduce the
	tokenisation process for raw privileged information.

	For static elements, the road map is represented as a set of dense points, as
	is the routing. The data format of dense points is redundant for our planner,
	since a set of dense points can be approximated through several control points~\cite{gao2020vectornet}.
	To reduce redundancy, we used the Ramer-Douglas-Peucker algorithm~\cite{ramer1972iterative,douglas1973algorithms}
	to approximate the dense points into a set of sparse control points. A rectangle
	with a certain width and height is then extended based on the location of the sparse
	control points. Segments, namely, routing segments
	$\mathcal{S}_{rt}\in\mathbb{R}^{N_{rt} \times 6}$ and road edge segments
	$\mathcal{S}_{rd}\in\mathbb{R}^{N_{rd} \times 6}$, are the collections of rectangles
	for static road elements. $N$ is the maximum number of rectangles within
	segments, and 6 attributes for each rectangle are $[x,y,w,h,yaw,id]$, where
	$id$ is the sequential numerical index for that rectangle. The extended width for
	$\mathcal{S}_{rt}$ and $\mathcal{S}_{rd}$ is set to the ego agent's width and $0
	.5m$ respectively.

	For non-ego agents, we tokenised their properties into 6 attributes
	$\mathcal{S}_{nego}\in\mathbb{R}^{N_{nego} \times 6}$ similarly to the static
	ones, except we replace the $id$ with $speed$ for each non-ego agent. For the
	ego agent $\mathcal{S}_{ego}\in\mathbb{R}^{1 \times 6}$, the attributes are aligned
	with $\mathcal{S}_{nego}$.

	It should be noted that the tokenisation for $\mathcal{S}_{rd}$ and $\mathcal{S}
	_{veh}$ is only performed within the Field of View (FOV) under the ego agent coordinate
	system, with a certain width $w_{f}$ and height $h_{f}$. After tokenisation,
	the stack for each segment is then represented as observation
	$\mathcal{O}=[\mathcal{S}_{rt},\mathcal{S}_{rd},\mathcal{S}_{nego},\mathcal{S}_{ego}]$
	for the planner. To distinguish each segment, we add an indicator to each one,
	thus,
	$\mathcal{O}\in \mathbb{R}^{(N_{rd} + N_{rt} + N_{nego}+1) \times (6+1)}$.

	The control signal $\mathcal{A}$ from the MLP is determined by the action
	space: $\mathcal{A}\in \mathbb{R}^{2}$ (\emph{i.e.,} $[steer, acc]$) if we use
	the bicycle model and $\mathcal{A}\in \mathbb{R}^{3}$ for waypoints (\emph{i.e.,}
	$[x, y, yaw]$).

	\paragraph{Training.}
	The training scheme for IL and RL is presented in Fig.~\ref{fig.pipeline},
	highlighted by different colours. For IL, the network is directly supervised
	by the actions from the expert log,
	\begin{equation}
		\mathcal{L}_{IL}= \left \| \mathcal{A}_{gt}- \mathcal{A}\right \|_{1}. \label{eq.bc}
	\end{equation}
	$\left \| \cdot \right \|_{1}$ represents the $l_{1}$ norm. For RL, inspired by~\cite{roach},
	we use a Beta distribution to parameterise the continuous bicycle action space.
	We use PPO (Proximal Policy Optimization) with clipping to train the network.
	$\mathcal{L}_{RL}$ is the clipped policy gradient loss with advantages
	estimated using Generalised Advantage Estimation~\cite{schulman2015high}. $\mathcal{L}
	_{value}$ is used to train the value network that is used to compute the advantage
	estimate,
	\begin{equation}
		\mathcal{L}_{value}= \left \| \mathcal{R}- \text{value}\right \|_{2}.
	\end{equation}
	$\mathcal{R}$ is the reward provided by the simulator. The "value" is the output
	of the value network, and $\left \| \cdot \right \|_{2}$ represents the
	$l_{2}$ norm.

	\paragraph{Reward Shaping.}
	Building on the existing reward described in~\cite{waymax}, the reward we used
	for RL consists of $R_{speed}$, $R_{offroad}$, $R_{wrongway}$, and $R_{collision}$.
	$R_{offroad}$ and $R_{collision}$ are consistent with the descriptions in~\cite{waymax}.
	The knowledge learned by the encoder from IL can be easily forgotten during RL
	iterations~\cite{mccloskey1989catastrophic}; therefore, we shape two additional
	rewards during the training process.

	$R_{speed}$ is a dense reward, representing the $l_{1}$ difference between the
	speed in the expert log and the actual speed during rollout.

	$R_{wrongway}$ is a binary reward, judging whether the ego agent is driving the
	wrong way. A trajectory is considered to be the wrong way if:
	\begin{itemize}
		\item The difference between the current orientation angle of the ego agent and
			the orientation angle of the closest point on its expert log trajectory is
			larger than $\Delta_{yaw}$.

		\item The Euclidean distance between the ego agent's current position and its
			nearest expert log trajectory point is greater than $\Delta_{dis}$.
	\end{itemize}

	The final reward $\mathcal{R}$ is the weighted sum of the rewards described
	above:
	\begin{equation}
		\mathcal{R}= w_{s}\times R_{speed}+ w_{o}\times R_{offroad}+ w_{c}\times R_{collision}
		+ w_{w}\times R_{wrongway}. \label{eq.reward}
	\end{equation}

	\begin{figure}[t]
		\centering
		\includegraphics[width=0.9\linewidth]{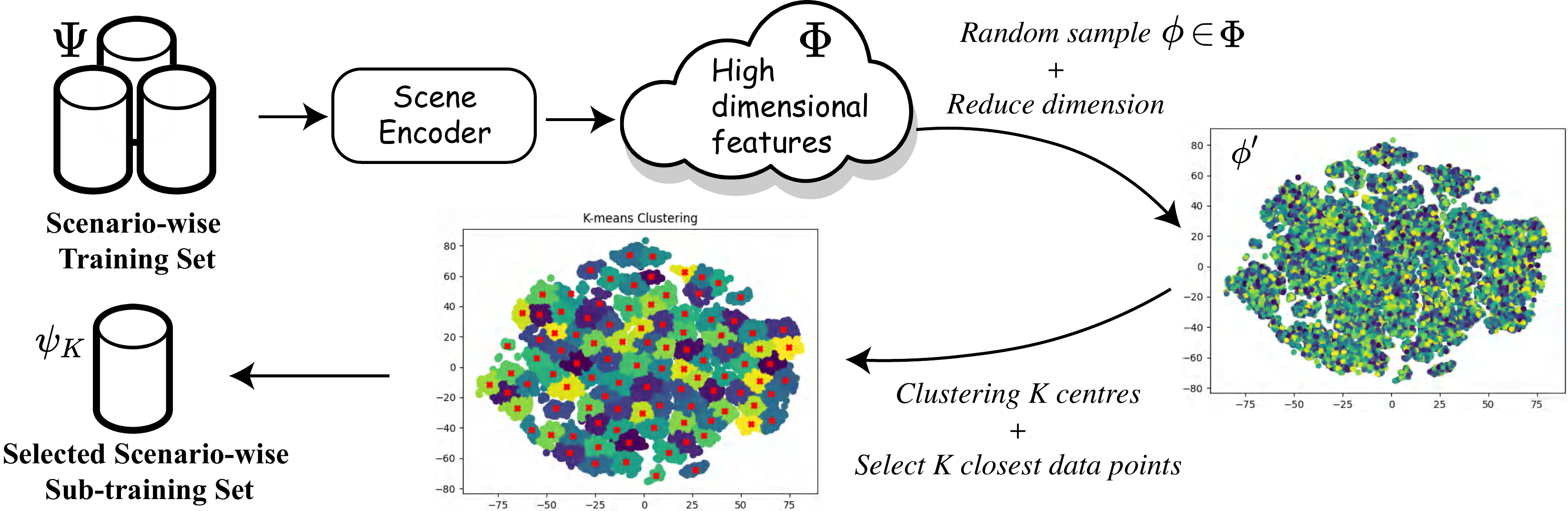}
		\caption{Illustration of SNE-Sampling. Scene encoder is pre-trained from EasyChauffeur.
		}
		\label{fig.kmeans}
	\end{figure}

	\subsection{Stochastic Neighbour Embedding Based Sampling}
	\label{sec.sampling}
	\paragraph{Motivation.}
	Previous enhancements in the literature~\cite{pdm,gameformer,hu2023imitation}
	have predominantly focused on network design and complex engineering
	approaches. However, our research demonstrates the equally crucial role of
	data. Our study revealed that training EasyChauffeur with only 10\% of the
	available training data in IL yielded performance comparable to using the full
	training set. This observation is in line with findings from~\cite{bronstein2023embedding}.

	Furthermore, in the context of limited data, we made the surprising discovery that
	RL exhibits superior data efficiency compared to IL. With this insight, our goal
	was to design a sampling procedure that can extract a sufficiently representative
	subset for RL training.

	Conventional random sampling techniques often assume a Gaussian distribution.
	While this assumption is reasonable if the entire training set follows a
	Gaussian distribution, it may not be optimal for better performance, as
	Gaussian distributions do not always represent the characteristics of neural
	networks well. Therefore, instead of this naive selection mechanism, we identify
	desired data points in the latent space of the networks.

	\paragraph{Procedure.}
	The aim of Stochastic Neighbour Embedding-based Sampling (SNE-Sampling) is to
	derive a representative subset containing $K$ scenarios, denoted as $\psi_{K}\subset
	\Psi$, from the entire training set $\Psi$. The complete SNE-Sampling procedure
	is illustrated in Fig.~\ref{fig.kmeans}.

	We begin with the scenario-wise training set, passing it through a scene encoder
	trained using IL to produce a set of high-dimensional features, denoted as
	$\Phi$. To reduce computational complexity, we initially sample a subset
	$\phi\subset\Phi$ at random, ensuring that the number of scenarios within $\phi$
	is significantly larger than $K$. We then apply t-SNE~\cite{van2008visualizing}
	to reduce the dimensionality of $\phi$, resulting in $\phi'$. Conceptually, a mapping
	function $f:\phi'\to\Psi$ exists such that each data point $p'_{i} \in \phi'$ can
	be associated with its corresponding origin in $\Psi$.

	To extract the subset $\psi_{K}$, we perform k-means clustering on $\phi'$ to
	obtain $K$ cluster centres. We then select the closest data point $p_{i} \in \phi
	'$ to each cluster centre and identify its corresponding mapping $f(p_{i}') \in
	\Psi$. Consequently, the scenario-wise subset $\psi_{K}$ comprises the collection
	of $(f(p_{1}'),...,f(p_{K}'))$.

	\begin{figure}[t]
		\centering
		\includegraphics[width=0.9\linewidth]{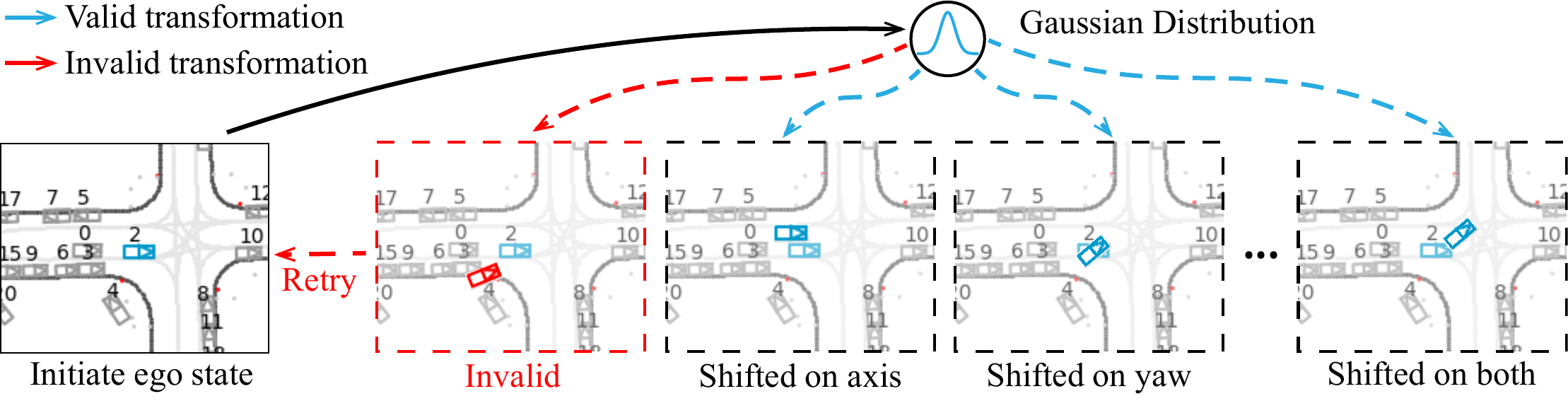}
		\caption{Diagram of Ego-Shifting. We perform a transformation on the ego agent
		on the x-y plane and the yaw axis; each transformed value is generated
		through a Gaussian distribution. }
		\label{fig.shift_demo}
	\end{figure}

	\subsection{Ego-Shifting for Robust Evaluation}
	\paragraph{Motivation.}
	Conventional supervised learning operates under the assumption that the
	training and test set distributions are statistically similar. However, this
	assumption does not hold in real-world applications when the training set
	lacks diversity. This limitation is particularly pronounced in \textit{close-loop} evaluation
	scenarios, as it is impractical to cover all possible routes from point A to B.
	Moreover, many \textit{close-loop} benchmarks presuppose that the ego agent's initial
	position is perfectly aligned, which fails to reflect real-world variances. In
	practice, ego agents may deviate from the intended route due to actions such as
	overtaking, pulling over, or imperfect localisation.

	\paragraph{Procedure.}
	Inspired by the above insight, we propose a novel evaluation perspective that can
	be applied to any existing \textit{close-loop} evaluation protocol.
	The Ego-Shifting process is depicted in Fig.~\ref{fig.shift_demo}. The
	transformation process involves two aspects: geometric and coordinate transformations.

	Geometrically, a maximum shifting value is defined for both the x-y plane and
	the yaw axis to regulate the transformation magnitude for each given initial
	ego state. The actual transformed values are generated using a Gaussian distribution.
	Validated transformations can occur along the axes (referred to as "shifted on
	axis"), along the yaw (referred to as "shifted on yaw"), or a combination of
	both (referred to as "shifted on both"). If a transformation is invalid, such
	as causing off-road incidents or collisions, we retry the process until the maximum
	number of retries is reached; otherwise, the initial state of the ego car is maintained.

	Coordinately, all observations are shifted accordingly after the
	transformation is validated. In other words, we ensure that the observations
	are consistent with the ego agent's coordinate system both before and after
	the transformation process.

	\section{Experiments}
	\subsection{Evaluation Metrics \& Settings}
	\label{sec.evaluation} All experiments are conducted under \textit{close-loop}
	evaluation following~\cite{waymax}. The descriptions of the evaluation metrics
	are listed below.

	\begin{itemize}
		\item \textbf{Off-road Rate (OR).} Off-road is a binary metric that
			describes whether a vehicle is on the road edge. A vehicle is considered off-road
			when overlap with the road edge is detected, and vice versa. 'Rate' refers
			to the percentage of episodes in which the metric is flagged at any timestep.
			The descriptions below are consistent.

		\item \textbf{Collision Rate (CR).} Collision is a binary metric that
			determines whether a vehicle is colliding with another object within the
			scene. When assessing each pair of objects, if their bounding boxes
			overlap in a top-down view in 2D during the same timestep, they are classified
			as being in collision.

		\item \textbf{Progress Ratio (PR).} The progress ratio describes the
			percentage of the route that a vehicle has completed compared to the log
			trajectory. Since the future drivable area is not disclosed in WOMD, we
			exclude it from our calculation; therefore, the maximum value for PR for
			our method is 100.

		\item \textbf{Arrival Rate (AR).} Arrival is a binary metric that determines
			whether a vehicle has arrived at the target point safely. A vehicle is considered
			to have arrived if it achieves more than a 90\% progress ratio without
			collision or going off-road.

		\item \textbf{Non-reactive / Reactive.} Indicates whether non-ego agents are
			controlled by replayed logs or by IDM~\cite{idm} during \textit{close-loop}
			evaluation. The implementation of IDM directly follows Waymax.
	\end{itemize}

	\subsection{Dataset}
	All the experiments are conducted on WOMD (v1.1.0). Results are reported on
	the validation split. The training of PPO and \textit{close-loop} evaluation
	utilise recent released simulator Waymax~\cite{waymax}, where data is also from
	WOMD. The sequence length for each scenario is 8 seconds recorded with 10 Hz, agents
	are controlled under the frequency of 10 Hz, maximum number of agents within a
	scenario is set to 128. Total scenario for training is 487,002 and 44,096 for
	validation. Since no drivable futures are provided by WOMD at this moment, the
	simulation will be terminated when ego agent considered arrive safely (description
	on \textbf{Evaluation Metrics \& Settings} Arrival Rate part) during \textit{close-loop}
	evaluation.

	\begin{table}[tp]
		\small
		\centering
            \renewcommand\arraystretch{1.1}
		\label{Tab.1}
		\begin{tabular}{c|c|c|c|cccc}
			\whline{0.8pt} \textbf{Model}       & \textbf{Training Policy} & \textbf{Action Space}      & \textbf{Routing}       & \textbf{AR$\uparrow$} & \textbf{OR$\downarrow$} & \textbf{CR$\downarrow$} & \textbf{PR*$\uparrow$} \\
			\whline{0.8pt} WF.~\cite{wayformer} & -                        & \multirow{2}{*}{Waypoints} & \multirow{2}{*}{LT+DF} & -                     & 7.89                    & 10.68                   & 123.58                 \\
			WM.~\cite{waymax}                   & IL                       &                            &                        & -                     & 4.14                    & 5.83                    & 79.58                  \\
			\hline
			\textbf{Ours}                       & IL                       & Waypoints                  & LT                     & 85.96                 & \textbf{2.80}           & \textbf{2.93}           & 95.77                  \\
			\whline{0.8pt}                      & IL                       & Bicycle                    & \multirow{2}{*}{LT+DF} & -                     & 13.59                   & 11.20                   & 137.11                 \\
			\multirow{-2}{*}{WM.~\cite{waymax}} & RL-DQN                   & Bicycle(discrete)          &                        & -                     & 4.31                    & 4.91                    & 215.26                 \\
			\hline
			                                    & IL                       &                            &                        & 73.60                 & 10.67                   & 4.92                    & 90.34                  \\
			\multirow{-2}{*}{\textbf{Ours}}     & RL-PPO                   & \multirow{-2}{*}{Bicycle}  & \multirow{-2}{*}{LT}   & \textbf{89.56}        & \textbf{2.16}           & \textbf{4.43}           & 98.00                  \\
			\whline{0.8pt}
		\end{tabular}
            \vspace{2mm}
		\caption{Comparison with previous planning models presented on Waymax~\cite{waymax}
		(WM.) and adaptation from Wayformer~\cite{wayformer} (WF.). `LT' and `DF' under
		\textbf{Routing} shorts for Logged Trajectory and Drivable Futures (unavailable),
		respectively. We implement our method for continuous action space waypoints
		and bicycle. \textbf{PR*}: \textit{it can not be compared directly for the
		unavailability of Driviable Futures}. All evaluated under
		the \textbf{Reactive} setting. }
	\end{table}

	\subsection{Implementation Details}
	\paragraph{Network Structure.}
	For both RL and IL, we choose BERT-mini~\cite{bhargava2021generalization} as transformer
	scene encoder. For IL, we use a 1-layer MLP with hidden dimension of 256. For RL
	we use a 2-layer MLP with hidden dimension of 64 for policy head and value head.
	To be notice that EasyChauffeur only generate control signal for next timestep
	rather than a sequence. This simplification is aimed at attenuating the impact
	caused by the model and emphasising the significant improvement brought about by
	changes in the training strategy.

	\paragraph{Preprocessing for Observation.}
	Ramer-Douglas-Peucker algorithm is used to sample dense points into sparse. The
	hyper-parameters for Ramer-Douglas-Peucker follow~\cite{plant} and can be found
	in supplemental materials. Instead of generating observation online during the
	training of IL, we will collect observation from 500 scenarios simultaneously and
	dump the concatenation of it for every simulated timesteps. Dynamic models to
	generate the ground truth control signal is aligned with the description on~\cite{waymax}.
	The observation has been transformed into ego coordinate system, thus only
	$speed$ in $\mathcal{S}_{ego}$ is a variance while others, namely,
	$[x,y,w,h,yaw]$ remain constant. For waypoints as action space, we do not fed
	ego agent since the \textit{speed} deteriorates performance~\cite{cheng2023rethinking},
	while the determinate of acceleration at next state necessitates it's current
	speed under bicycle model according to Newton's laws of motion.

	\paragraph{Hyperparameters.}
	The width $w_{f}$ and height $h_{f}$ of FOV is set to 80m and 20m respectively.
	Threshold $\Delta_{yaw}$ and $\Delta_{dis}$ for $R_{wrongway}$ is set to 1
	radians and 3.5m. Since all the weights on Eqn.~\ref{eq.reward} is binary except
	$w_{s}$ is continuous, we intuitively set the binary one as -1.0 and continuous
	one as 1.0. We select 100, 400, 3,150 as $K$ during SNE-Sampling and training
	for RL. Please refer to supplemental materials for detail setups and hyperparameters
	for IL and RL.

	\subsection{Effectiveness of EasyChauffeur}

	To demonstrate the efficacy of EasyChauffeur, in Tab.~\ref{Tab.1}, we present comprehensive
	comparison with the results reported in~\cite{waymax}. Consistent with~\cite{waymax},
	we implement our method using action space waypoints (3rd row) and bicycle (6th
	row). In comparison to the methods reported in the literature (2nd and 4th rows),
	our method exhibits superior performance on metrics that can be fairly compared,
	such as Reactive-OR and Reactive-CR. These results establish the strength of our
	method as a pivotal reference point for further comparisons.

	We do not implement our method under discrete action space for following reasons.
	First, the purpose of this paper is to provide a feasible \& effective baseline,
	which has been proven under two different action space. Second, the detail for
	discretization under different action space is neither reported on~\cite{waymax,wayformer}
	nor within the scope of this paper, we leave this open for future researchers.

	The selection of PPO over Deep Q-Network (DQN) in the Bicycle model is based on
	several primary considerations. Firstly, employing waypoint action space on Waymax
	is unrealistic due to the direct updating of the absolute position for the next
	simulated state. Consequently, not all transitions are kinematically compatible,
	and applying RL under waypoint action space may compromise physical
	constraints, leading to intricate reward shaping. Secondly, PPO supports both discrete
	and continuous action spaces. Furthermore, as our aim is to propose a baseline
	that is easy to reimplement, PPO generally exhibits better training stability
	compared to DQN. DQN is known to be sensitive to hyperparameter settings and can
	suffer from issues such as Q-value overestimation, which can impact training
	performance~\cite{van2016deep}

	\subsection{Insights from Training with Limited Data}
	Tab.~\ref{Tab.rand} presents the results of an ablation study on the volume of
	training data for various training strategies. The entire dataset is randomly sampled
	from the complete training set. Notably, an intriguing phenomenon emerges during
	the ablation study of IL: the network achieves comparable performance with
	only 10\% of the data, as indicated by Non-reactive-AR and Reactive-AR. This premature
	saturation, accompanied by a low upper bound, suggests that the majority of driving
	scenarios are relatively simple to replicate, while challenging corner cases
	remain difficult to capture.

	This observation is further supported by experiments involving the same
	network trained with the PPO strategy, as depicted in the last three rows of Tab.~\ref{Tab.rand}.
	Clear evidence indicates that the upper performance bound of the same network on
	the same data is significantly improved, although premature saturation
	persists. This finding suggests that some of the corner cases have been explored
	through numerous rollouts.

	Based on these observations, we contend that instead of utilising the entire
	biased training set, selecting a representative subset has the potential to enhance
	performance. We chose 3,150 as the maximum number of scenario as this is the
	maximum number that our GPUs can perform rollout in parallel, which can be further
	engineeringly optimised.

	\begin{table}[tp]
		\small
		\centering
            \renewcommand\arraystretch{1.1}
		\label{Tab.rand}
		\begin{tabular}{c|c|c|cccc|cccc}
			\whline{1pt} \multirow{2}{*}{\textbf{\begin{tabular}[c]{@{}c@{}}Training \\ policy\end{tabular}}} & \multirow{2}{*}{\textbf{\begin{tabular}[c]{@{}c@{}}Scenario\\ number\end{tabular}}} & \multirow{2}{*}{\textbf{\begin{tabular}[c]{@{}c@{}}Volume of \\ training data\end{tabular}}} & \multicolumn{4}{c|}{\textbf{Non-reactive}} & \multicolumn{4}{c}{\textbf{Reactive}} \\
			\cline{4-11}                                                                                      &                                                                                     &                                                                                              & \textbf{AR$\uparrow$}                      & \textbf{OR$\downarrow$}              & \textbf{CR$\downarrow$} & \textbf{PR$\uparrow$} & \textbf{AR$\uparrow$} & \textbf{OR$\downarrow$} & \textbf{CR$\downarrow$} & \textbf{PR$\uparrow$} \\
			\whline{1pt}                                                                                      & 100                                                                                 & $\sim$0.02\%                                                                                 & 50.24                                      & 19.53                                & 24.62                   & 78.18                 & 52.22                 & 19.16                   & 8.50                    & 78.57                 \\
			                                                                                                  & 400                                                                                 & $\sim$0.08\%                                                                                 & 57.59                                      & 12.59                                & 18.25                   & 82.96                 & 58.71                 & 12.61                   & 6.97                    & 83.24                 \\
			                                                                                                  & 3,150                                                                               & $\sim$0.65\%                                                                                 & 60.19                                      & 15.73                                & 16.02                   & 86.04                 & 61.69                 & 15.81                   & 8.40                    & 86.29                 \\
			                                                                                                  & 48,700                                                                              & $\sim$10\%                                                                                   & 74.78                                      & 9.44                                 & 9.37                    & 89.57                 & 71.80                 & 9.57                    & 8.80                    & 89.52                 \\
			\multirow{-6}{*}{IL}                                                                              & 487,001                                                                             & 100\%                                                                                        & 72.48                                      & 10.82                                & 8.28                    & 89.93                 & 73.60                 & 10.67                   & 4.92                    & 90.34                 \\
			\hline
			\multirow{3}{*}{RL-PPO}                                                                           & 100                                                                                 & $\sim$0.02\%                                                                                 & 84.38                                      & 5.84                                 & 6.41                    & 97.72                 & 83.87                 & 5.91                    & 5.82                    & 97.79                 \\
			                                                                                                  & 400                                                                                 & $\sim$0.08\%                                                                                 & 85.35                                      & 2.47                                 & 3.63                    & 95.51                 & 85.30                 & 2.55                    & 4.24                    & 96.19                 \\
			                                                                                                  & 3,150                                                                               & $\sim$0.65\%                                                                                 & 86.76                                      & 1.91                                 & 3.11                    & 94.53                 & 86.29                 & 1.94                    & 3.69                    & 95.01                 \\
			\whline{1pt}
		\end{tabular}
            \vspace{2mm}
  		\caption{Ablation of the volume of training data for different training
		policies.}
	\end{table}

	\subsection{Effectiveness of SNE-Sampling}
	The efficacy of our SNE-Sampling technique is showcased in Tab.~\ref{Tab.kmeans},
	wherein the performance of various configurations is evaluated. Notably, all
	configurations exhibit enhancements in terms of AR, except for Non-reactive-AR,
	which shows a marginal degradation of -0.73 within acceptable limits. It is
	observed that as the magnitude of the scenario number escalates, the impact of
	SNE-Sampling becomes increasingly potent, manifesting in two discernible
	manners. Firstly, in comparison to random sampling, the augmentation in data volume
	yields greater performance improvements. Secondly, with each incremental increase
	in data volume, the advantage of SNE-Sampling over random sampling continues
	to expand.

	\begin{table}[tp]
		\label{Tab.kmeans}
		\centering
		\small
            \renewcommand\arraystretch{1.1}
		\setlength{\tabcolsep}{1.6mm}
		\begin{tabular}{c|cccc|cccc}
			\whline{1pt} \multirow{2}{*}{\textbf{\begin{tabular}[c]{@{}c@{}}\#Scenario \end{tabular}}} & \multicolumn{4}{c|}{\textbf{Non-reactive}} & \multicolumn{4}{c}{\textbf{Ractive}} \\
			\cline{2-9}                                                                                       & \textbf{AR$\uparrow$}                      & \textbf{OR$\downarrow$}             & \textbf{CR$\downarrow$} & \textbf{PR$\uparrow$} & \textbf{AR$\uparrow$} & \textbf{OR$\downarrow$} & \textbf{CR$\downarrow$} & \textbf{PR$\uparrow$} \\
			\whline{1pt} 100                                                                                  & 84.38                                      & 5.84                                & 6.41                    & 97.72                 & 83.87                 & 5.91                    & 5.82                    & 97.79                 \\
			\multicolumn{1}{l|}{+SNE-Sampling}                                                                & $83.65_{(-0.73)}$                          & 5.23                                & 4.91                    & 96.30                 & $84.04_{(+0.17)}$     & 5.12                    & 5.62                    & 97.00                 \\
			\hline
			400                                                                                               & 85.35                                      & 2.47                                & 3.63                    & 95.51                 & 85.30                 & 2.55                    & 4.24                    & 96.19                 \\
			\multicolumn{1}{l|}{+SNE-Sampling}                                                                & $87.29_{(+1.94)}$                          & 2.76                                & 3.62                    & 96.82                 & $87.12_{(+1.82)}$     & 2.80                    & 4.95                    & 97.41                 \\
			\hline
			3,150                                                                                             & 86.76                                      & 1.91                                & 3.11                    & 94.53                 & 86.29                 & 1.94                    & 3.69                    & 95.01                 \\
			\multicolumn{1}{l|}{+SNE-Sampling}                                                                & $90.68_{{(+3.92)}}$                        & 2.15                                & 2.58                    & 97.51                 & $89.56_{(+3.27)}$     & 2.16                    & 4.43                    & 98.00                 \\
			\whline{1pt}
		\end{tabular}
            \vspace{2mm}
  		\caption{Effectiveness of SNE-Sampling. AR under both Non-reactive and Reactive
		generally experience improvement for three different scenario number settings.}
	\end{table}

	\begin{figure}[t]
            \centering
		\includegraphics[width=0.95\linewidth]{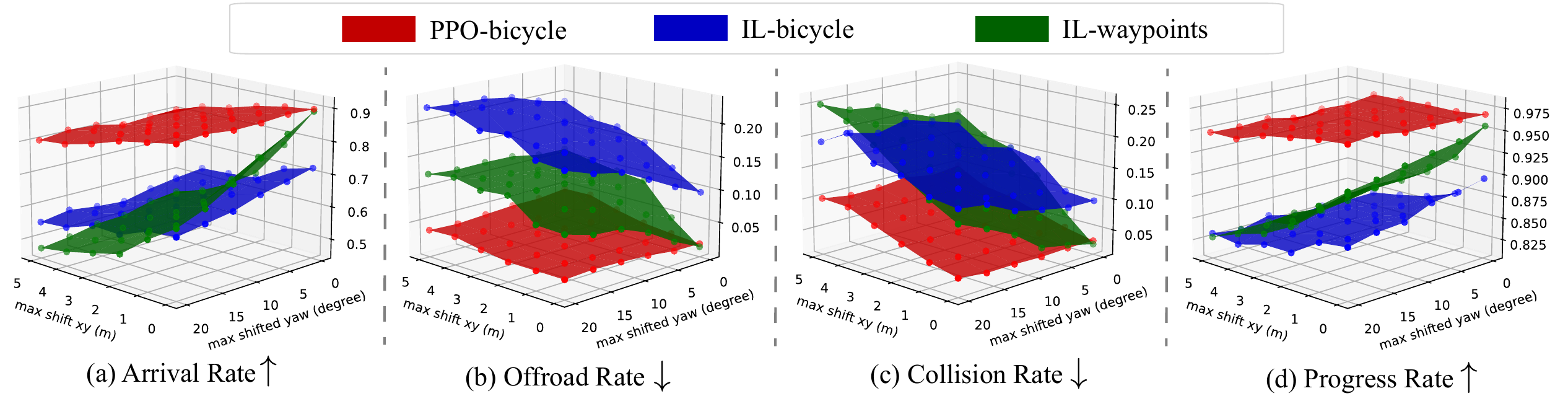}
		\caption{The performance of EasyChauffeur is evaluated under various
		training policies, Ego-Shifting properties, and action spaces.}
		\label{fig.shift_results}
	\end{figure}

	\subsection{Experiments on Ego-Shifting}
	\paragraph{Comparison under Different Ego-Shifting Properties. }
	To enhance the efficiency of evaluation, we conducted an experiment on Ego-Shifting
	within a subset comprising 1050 scenarios involving Non-reactive agents. We specifically
	selected six steps based on the maximum values of shifted yaw and shifted xy,
	namely $[0,4,8,15,20]$ degrees and $[0,1,2,3,4,5]$ meters, resulting in a total
	of 35 experiments. Subsequently, we plotted the trends of maximum shifting values
	against four evaluation metrics for EasyChauffeur, trained using PPO-bicycle, IL-bicycle,
	and IL-waypoints. The performance of EasyChauffeur under the aforementioned settings
	is depicted in Fig.~\ref{fig.shift_results}.

	Three key findings can be inferred from the results. Firstly, conventional IL
	methods exhibit significant variance in performance when the distribution of
	the validation set deviates from that of the training set. Notably, this
	variance increases as the degree of shifting intensifies. Secondly, positional
	changes are more sensitive to orientation angles. Illustratively, in the case
	of IL-waypoints (highlighted in green), the lowest arrival rate along the yaw axis
	(65.24\%) is considerably higher than that along the xy axis (52.86\%).
	Thirdly, even in the absence of any domain adaptation techniques, PPO-bicycle
	demonstrates superior long-term robustness compared to IL methods.

\begin{figure}[t]
		\centering
		\includegraphics[width=0.8\linewidth]{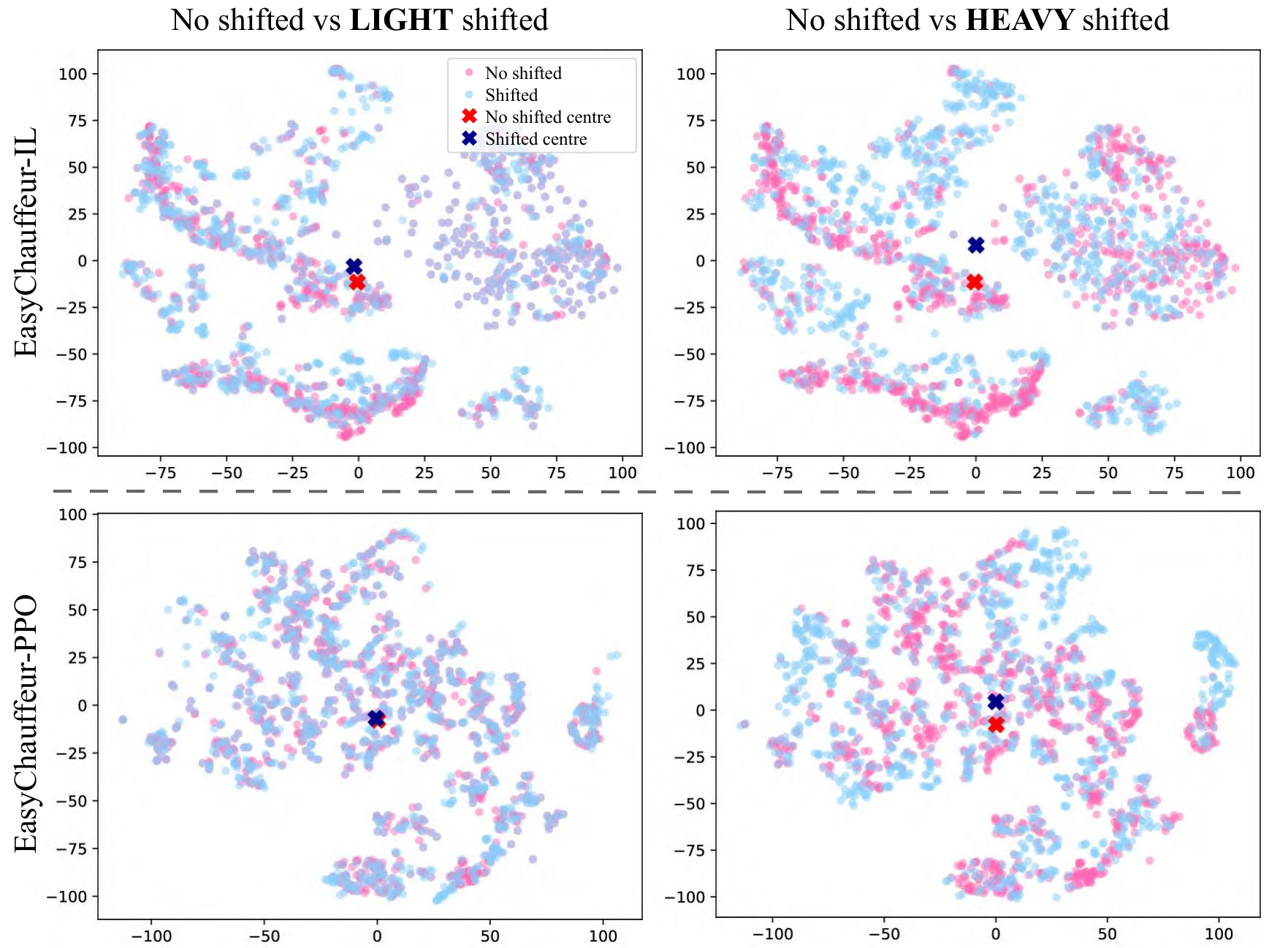}
		\caption{Visualisation of feature distribution before/after Ego-Shifting for
		EasyChauffeur-IL and EasyChauffeur-PPO.}
		\label{fig.shift_dis}
	\end{figure}

	\paragraph{Analysis of Ego-Shifting.}
	The concept of Ego-Shifting entails introducing out-of-distribution noise into
	a given distribution, thereby enabling the assessment of the anti-noise
	capabilities of different models. The visualisation of the feature distribution
	before and after performing Ego-Shifting for two models is presented in Fig.~\ref{fig.shift_dis}.

	In the visualisation, the pink scattering denotes features belonging to the original
	data, while the light blue scattering represents the shifted data. The red and
	blue `X' marks indicate the shifted centres for the data. Two degrees of
	shifting are visualised: LIGHT and HEAVY, corresponding to transformation
	pairs of (1m, 4 degrees) and (5m, 20 degrees), respectively.

	Analysing the first row of Fig.~\ref{fig.shift_dis}, it becomes evident that the
	Ego-Shifting method proposed effectively alters the distribution of the validation
	set. As the application of LIGHT to the HEAVY transformation progresses from
	left to right, the distance between the non-shifted centre and the shifted centre
	increases, and the overlap area of two decrease. This observation serves as confirmation
	of the effectiveness of the proposed Ego-Shifting mechanism in modifying the distribution.
	Furthermore, the second row of Fig.~\ref{fig.shift_dis} provides additional
	evidence highlighting the robustness of RL to environmental changes. Comparing
	the distances between the two centres and the overlap area of two clusters, it
	can be observed that the changes are minimal, indicating the ability of RL to adapt
	to the shifting environment.

	\section{Conculsion}
	In this paper, we illustrate that integrating reinforcement learning with minimal
	data can significantly enhance performance, facilitated by our novel SNE-Sampling
	technique. Furthermore, our Ego-Shifting evaluation method introduces a new
	perspective for \textit{close-loop} evaluation, effectively reducing the sim-to-real gap.
	We advocate for a comprehensive strategy that balances innovative training, efficient
	data utilization, and rigorous evaluation to advance autonomous driving
	technologies.

	\bibliographystyle{unsrtnat}
	\bibliography{references}
	\input{appendix}
\end{document}

%% file: appendix.tex
\clearpage
\appendix
\newcommand{\addFig}[2][]{\includegraphics[width=0.25\linewidth, #1]{figs/#2}}
\newcommand{\addscentext}[1]{\rotatebox{90}{#1}}
\newcommand{\STAB}[1]{\begin{tabular}{@{}c@{}}
	#1
\end{tabular}}

\section{Videos}
We provide videos on \textit{close-loop} evaluation, located in a folder named \texttt{supp-demonstrations}\footnote{Download
link: \url{https://drive.google.com/drive/folders/1LpAeBZJzXb7Srgq0fTY9KC6DJY6XjajD}}.
We categorise the videos by scenario into three sub-folders named \texttt{parkinglot},
\texttt{carriageways}, and \texttt{urban}. Within each sub-folder, you will find
cases introduced in Figs.~\ref{fig.supp-lot}, \ref{fig.supp-highway-a}, \ref{fig.supp-highway-b}, and
\ref{fig.supp-urban}, named row-wise accordingly. For example, \texttt{1.mp4}
refers to the first row in the figure. In addition to the mentioned cases, we
provide more examples under \texttt{examples}. The file structure is organised
as below:

\dirtree{%
.1 supp-demonstrations. .2 parkinglot. .3 examples. .4 .... .3 1.mp4. .3 ... .2 .... }

\begin{figure}[tp]
	\centering
	\includegraphics[width=0.9\linewidth]{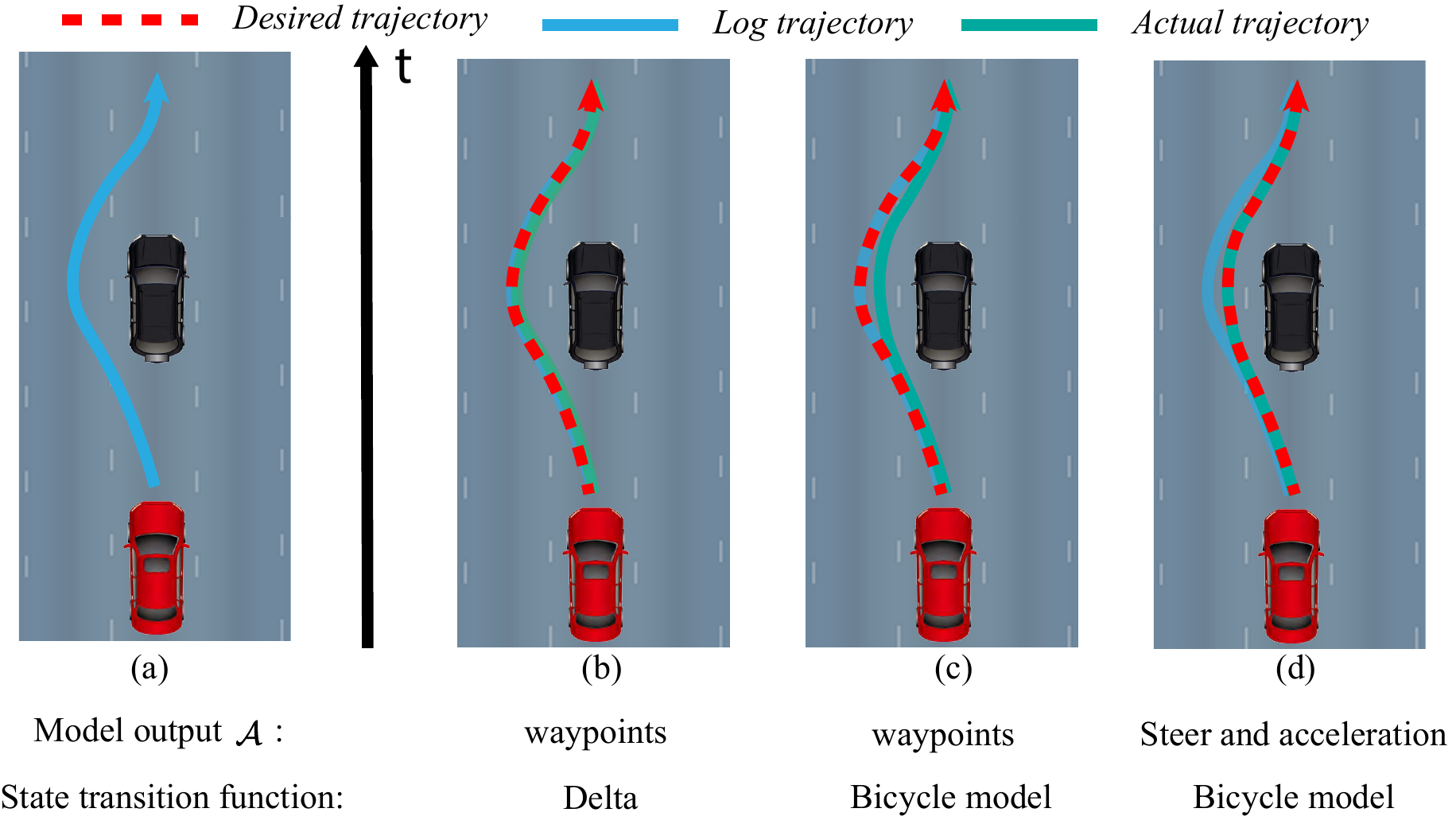}
	\caption{Illustration of different combinations of model output and state
	transition function.}
	\label{fig.dynamicmodel}
\end{figure}
\section{Dynamic Model}
Waymax ~\cite{waymax} provided two types of transition function to calculate agents'
next state: `Delta' and `Bicycle model'. For `Delta' state transition function, the
transition from $t$ to $t+1$ is directly determined by the next state's coordinate
that fed into simulator, regardless it's kinematics feasibility, thus `Delta' may
not consistent with reality application (may occurs teleport or spin in this
case). On the other hand, for `Bicycle model' as transition function, the
simulator will calculate agents' next state by steer and acceleration, which
ensure that every state transitions are satisfy kinetic constraints.

Different combinations of model output $\mathcal{A}$ and state transition
function are demonstrated in Fig.~\ref{fig.dynamicmodel}. The desired trajectory
is the trajectory that calculated by the set of model's output. As Fig.~\ref{fig.dynamicmodel}(b)
shows, there can be a gap between desired trajectory and actual trajectory if
the action provided by model is not aligned with the state transition function, since
a set of waypoints need to first fit by steer and acceleration before transited (also
discussed on~\cite{cheng2023rethinking}). Thus, when referring waypoints as action
space we use combination on Fig.~\ref{fig.dynamicmodel}(a) and bicycle as Fig.~\ref{fig.dynamicmodel}(c).

\noindent
\textbf{Waypoints.} In this case, the model's output is $\mathcal{A}=(\Delta x, \Delta
y, \Delta yaw)$. Given the current agent's state $(x,y,yaw,v_{x},v_{y})$, the next
state $(x',y',yaw',v_{x}',v_{y}')$ is updated as:

\begin{equation}
	\begin{split}
		&x' = x + \Delta x \\&y' = y + \Delta y \\&yaw' = yaw + \Delta yaw\\&v_{x}' =
		(x'-x)\cdot f\\&v_{y}' = (y'-y)\cdot f.\\
	\end{split}
\end{equation}
$f$ is the control frequency, which is 10Hz as mentioned in the paper. $v_{x}$ and
$v_{y}$ is the velocity alone x-axis and y-axis.

\noindent
\textbf{Bicycle.} In this case, the model's output is $\mathcal{A}=(a,s)$, $a$
and $s$ are acceleration and steer for the agent, respectively. Next state is
updated according to:
\begin{equation}
	\begin{split}
		&x' = x+\frac{v_{x}}{f}+ \frac{a \cdot \cos(yaw)}{2f^{2}}\\&y' = y + \frac{v_{y}}{f}
		+ \frac{a \cdot \sin(yaw)}{2f^{2}}\\&yaw' = yaw + s \cdot (\frac{\sqrt{v_{x}^{2}+v_{y}^{2}}}{f}
		+ \frac{a}{2f^{2}})\\&v' = \sqrt{v_{x}^{2}+v_{y}^{2}}+ a/f \\&v_{x}' = v' \cdot
		\cos(yaw)\\&v_{y}' = v' \cdot \sin(yaw).\\
	\end{split}
\end{equation}

\section{Additional Hyperparameters}
\noindent
\textbf{Imitation Learning.} All experiments are conducted on 8 NVIDIA A800 GPUs.
Since $\mathcal{A}_{gt}\in \mathbf{R}^{n}$, where $n$ is the dimension of action
space, Eqn.~\ref{eq.bc} should be expressed as:
\begin{equation}
	\mathcal{L}_{IL}= w_{acc}\left \| a_{gt}- a \right \|_{1} + w_{steer}\left \| s
	_{gt}- s \right \|_{1}, \\ \label{eq.supp-bi}
\end{equation}
for bicycle as dynamic model, or
\begin{equation}
	\mathcal{L}_{IL}= w_{x}\left \| \Delta x_{gt}- \Delta
	x \right \|_{1} + w_{y}\left \| \Delta y_{gt}- \Delta y \right \|_{1} + w_{yaw}
	\left \| \Delta yaw_{gt}- \Delta yaw \right \|_{1},\label{eq.supp-wpts}
\end{equation}
for waypoints as dynamic model.
\begin{table}[]
	\label{tb.supp-il}
	\centering
        \renewcommand\arraystretch{1.1}
	\begin{tabular}{c|cc}
		\whline{1pt} \multirow{2}{*}{\textbf{Parameters}}   & \multicolumn{2}{c}{\textbf{Dynamic Model}} \\
		\cline{2-3}                                         & \multicolumn{1}{c|}{bicycle}              & waypoints \\
		\whline{1pt} learning rate                          & \multicolumn{2}{c}{$1\times10^{-4}$}       \\
		\hline
		decay schedule                                      & \multicolumn{2}{c}{None}                   \\
		\hline
		optimiser                                           & \multicolumn{2}{c}{Adam}                   \\
		\hline
		epochs                                              & \multicolumn{2}{c}{5}                      \\
		\hline
		scenarios per-batch                                 & \multicolumn{2}{c}{500}                    \\
		\hline
		batch size                                          & \multicolumn{2}{c}{6}                      \\
		\hline
		$w_{acc}$ \textbackslash{} $w_{ x}$                 & \multicolumn{2}{c}{1}                      \\
		\hline
		$w_{steer}$ \textbackslash{} $w_{ y}$ \& $w_{ yaw}$ & \multicolumn{1}{c|}{5}                    & 50        \\
		\whline{1pt}
	\end{tabular}
        \vspace{2mm}
 	\caption{Hyperparameters for IL. Batch size and learning rate are reported for
	per-GPU.}
\end{table}

\noindent
\textbf{Proximal Policy Optimization.} The implementation of PPO is based on stable-baseline3\footnote{\url{https://github.com/DLR-RM/stable-baselines3}}.
Practically, we follow~\cite{roach} to introduce an entropy loss
$\mathcal{L}_{ent}$ to encourage exploration. Therefore the objective for PPO can
be expressed as:
\begin{equation}
	\mathcal{L}_{ppo}= \mathcal{L}_{RL}+ w_{ent}\mathcal{L}_{ent}+ w_{value}\mathcal{L}
	_{value}.
\end{equation}
During evaluation, we take the mean of Beta distribution $\mathcal{B}$ as the deterministic
output.

\begin{table}[ht]
	\centering
        \renewcommand\arraystretch{1.1}
	\label{tb.supp-rl}
	\begin{tabular}{lccc}
		\whline{1pt} \multicolumn{4}{l}{\textit{\textbf{Rollout}}}  \\
		\whline{1pt} rollout scenarios $K$                         & 100                                      & ~~400 & 3,150 \\
		rollout GPUs                                               & 1                                        & ~~2   & 7     \\
		total timesteps                                            & 6M                                       & ~~16M & 25M   \\
		action range ($acc/steer$)                                 & \multicolumn{3}{c}{$[-6,6]$/$[-0.3,0.3]$} \\
		\whline{1pt} \multicolumn{4}{l}{\textit{\textbf{Training}}} \\
		\whline{1pt} learning rate                                 & \multicolumn{3}{c}{$3 \times 10^{-4}$}    \\
		epochs                                                     & \multicolumn{3}{c}{1}                     \\
		optimser                                                   & \multicolumn{3}{c}{Adam}                  \\
		batch size                                                 & \multicolumn{3}{c}{2,500}                 \\
		scenarios per-batch                                        & \multicolumn{3}{c}{1}                     \\
		$\gamma$ for GAE                                           & \multicolumn{3}{c}{0.99}                  \\
		$\lambda$ for GAE                                          & \multicolumn{3}{c}{0.9}                   \\
		max norm gradient clipping                                 & \multicolumn{3}{c}{0.5}                   \\
		clipping range for PPO                                     & \multicolumn{3}{c}{0.2}                   \\
		decay schedule                                             & \multicolumn{3}{c}{None}                  \\
		$w_{ent}$, $w_{value}$                                     & \multicolumn{3}{c}{1, 0.01}               \\
		$w_{s}$, $w_{o}$, $w_{c}$, $w_{w}$                         & \multicolumn{3}{c}{1, -1, -1, -1}         \\
		\whline{1pt}
	\end{tabular}
        \vspace{2mm}
 	\caption{Hyperparameters for PPO. We only use 1 GPU to update model's weights.
	The `epochs' under \textbf{\textit{Training}} means the times to iterate all collected
	data from rollout.}
\end{table}

\section{Visualisation of SNE-Sampling}
\begin{figure}
	\centering
	\includegraphics[width=\linewidth]{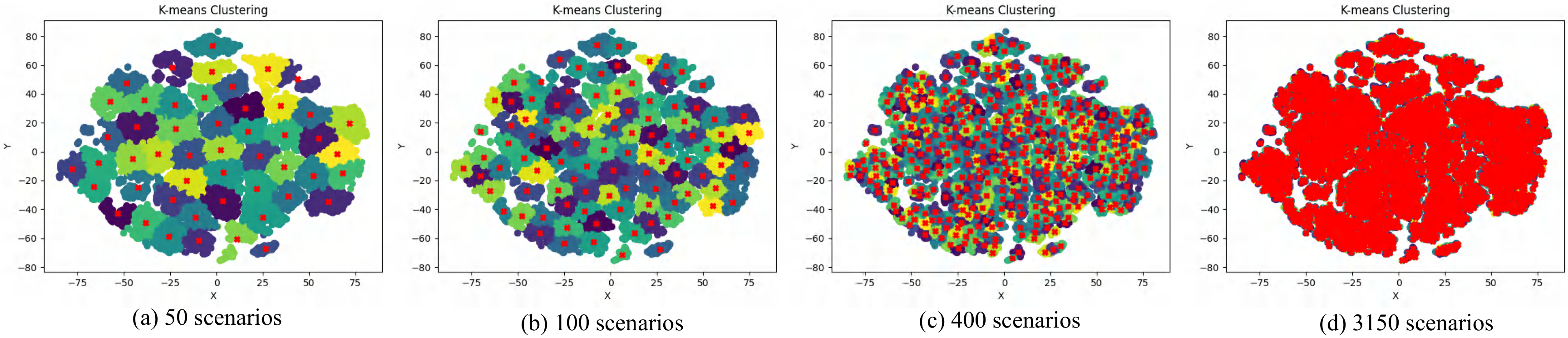}
	\caption{Visualisation of clustering centre (highlighted in red mark) for
	different $K$.}
	\label{fig.supp-kmeans}
\end{figure}

Fig.~\ref{fig.supp-kmeans} illustrates the clustering centre under different settings
of $K$, the number of scenarios. It is evident that with an increase in $K$ from
50 to 3,150, the coverage rate of the red mark becomes comprehensive. When selecting
100 scenarios, the selected data are not sufficiently representative to observe
a performance gain. However, with 400 scenarios, the improvement can be immediately
highlighted.

\section{Visualisation Results Across Varied Scenarios}
\label{sec.supp-vis}

\begin{figure*}[tp]
	\centering
	\resizebox{\linewidth}{!}{
	\begin{tabular}{ccccc}
		                                                                                        & \multicolumn{2}{c}{EasyChauffeur-IL}                                   & \multicolumn{2}{c}{EasyChauffeur-PPO}                       \\
		                                                                                        & w/o ego-shifting                                                       & w/ ego-shifting                                            & w/o ego-shifting                                           & w/ ego-shifting                                            \\
		\multirow{2}{*}[+3em]{\addscentext{Straight Routing}}                                   & \addFig[trim=100 100 100 100, clip]{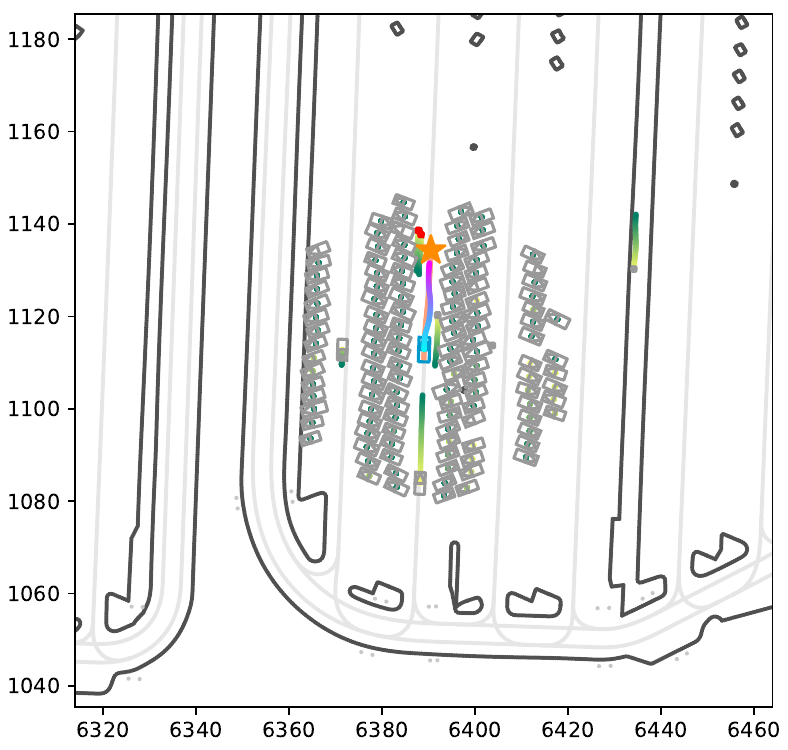}               & {\addFig[trim=100 100 100 100, clip]{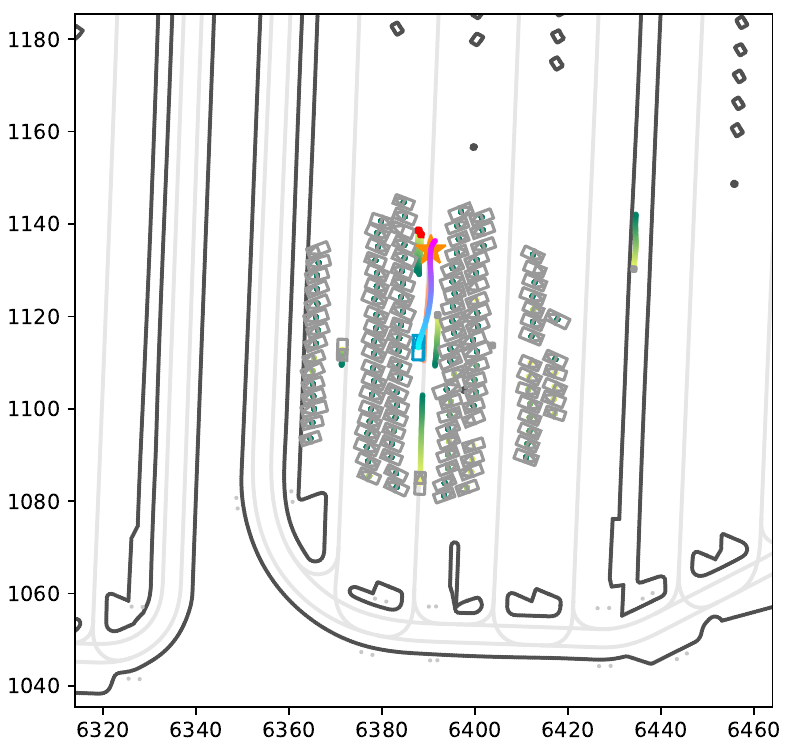}} & {\addFig[trim=100 100 100 100, clip]{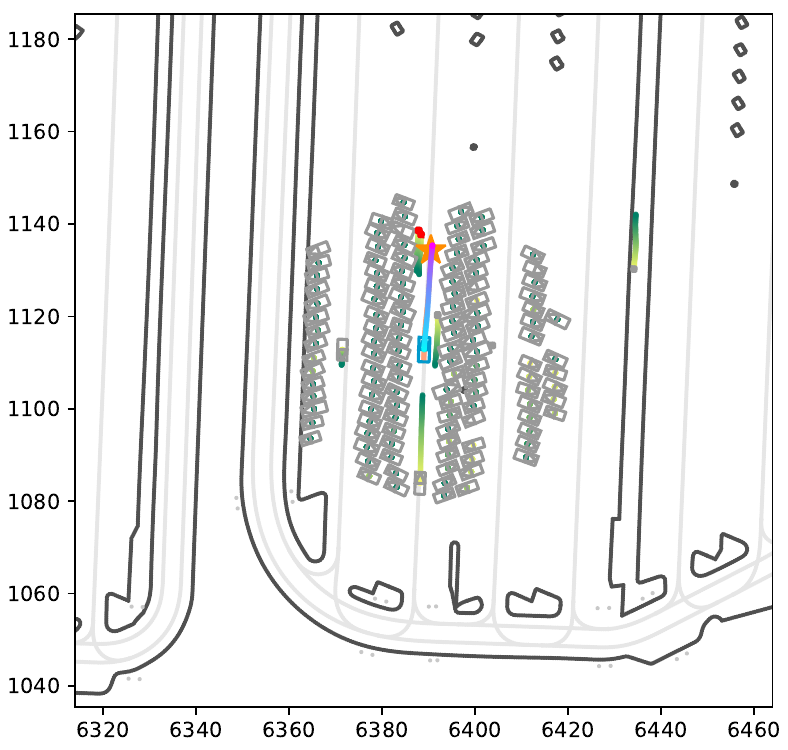}} & {\addFig[trim=100 100 100 100, clip]{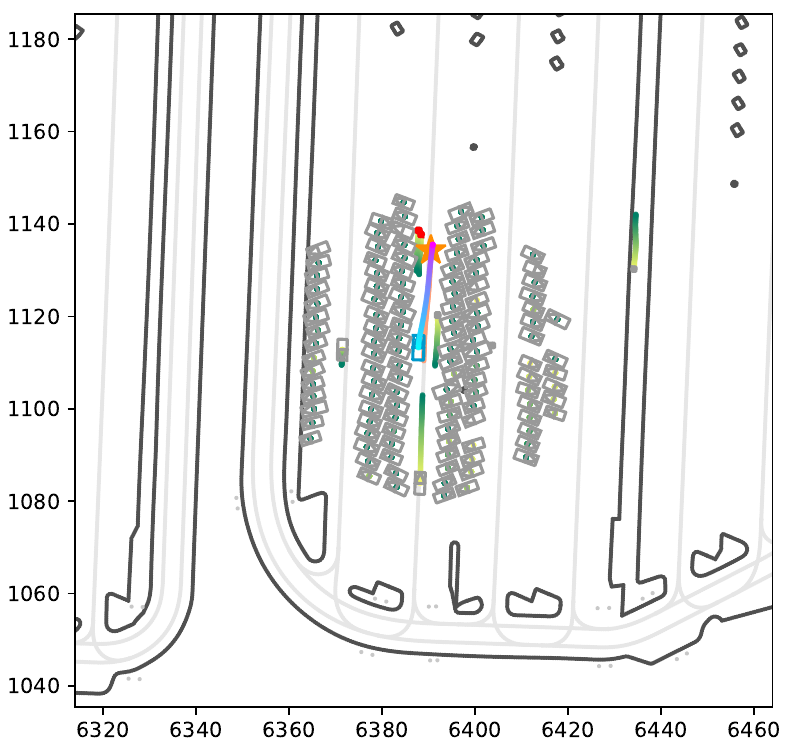}} \\
		                                                                                        & {\addFig[trim=100 100 100 100, clip]{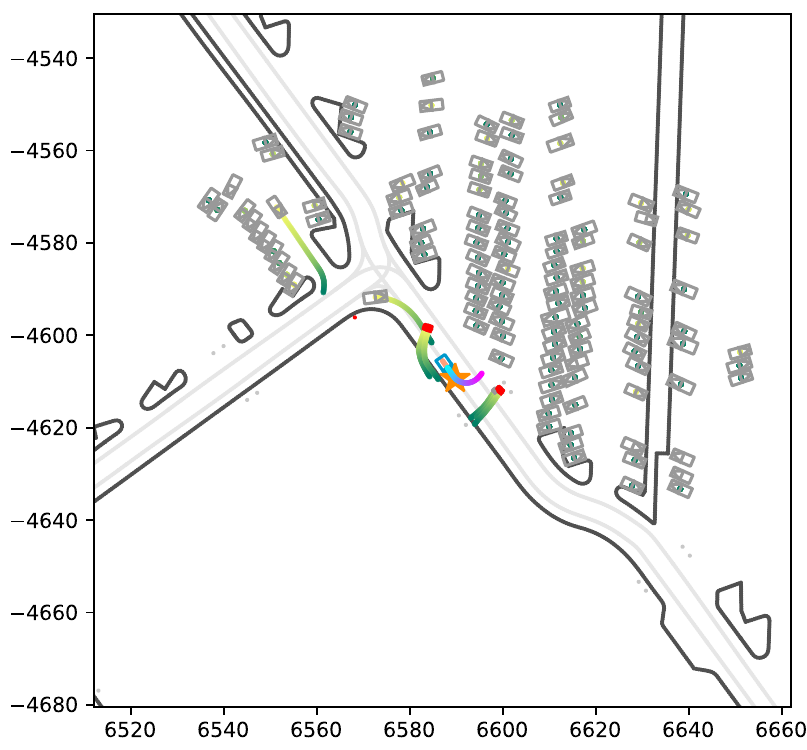}}             & {\addFig[trim=100 100 100 100, clip]{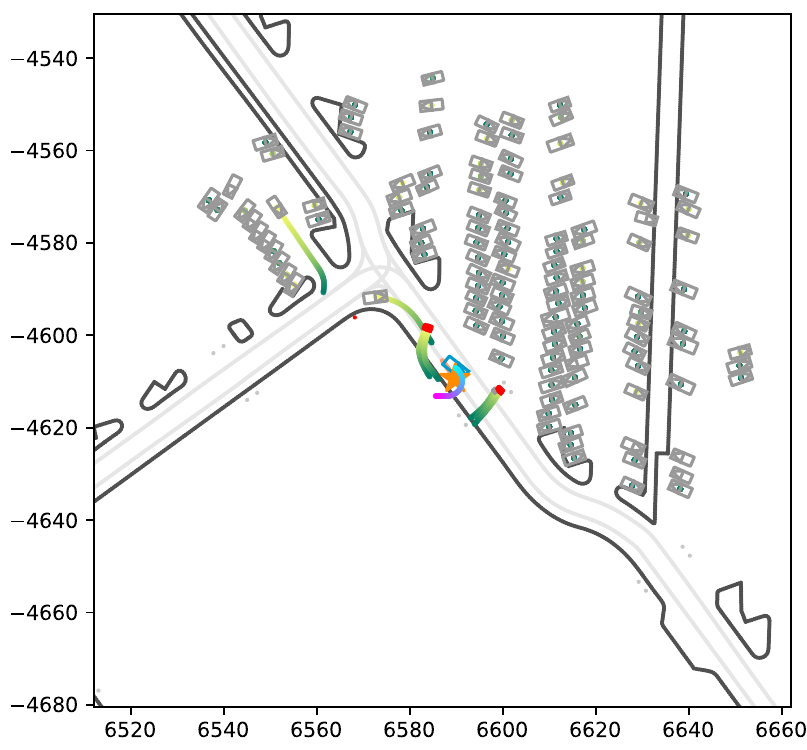}} & {\addFig[trim=100 100 100 100, clip]{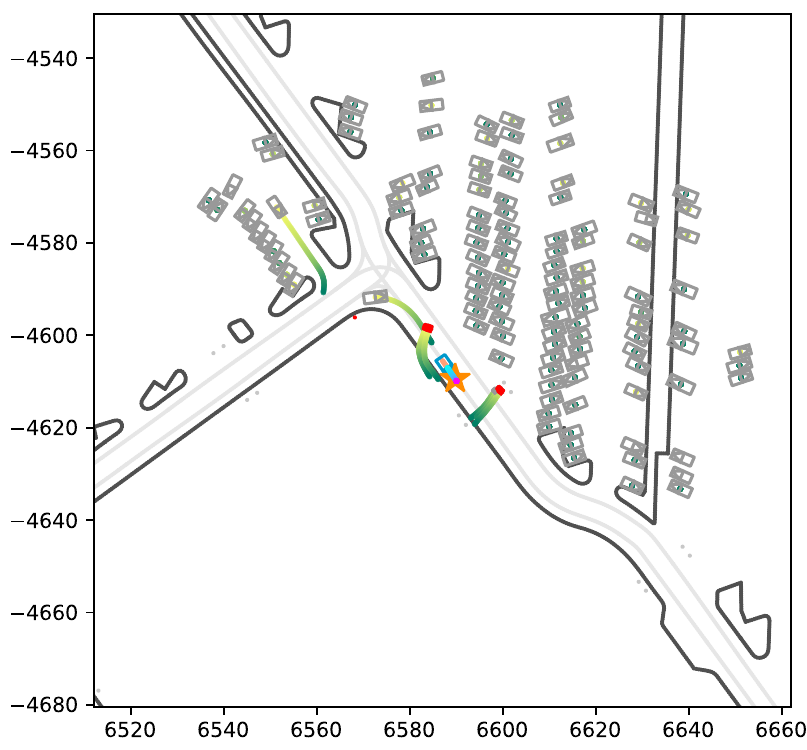}} & {\addFig[trim=100 100 100 100, clip]{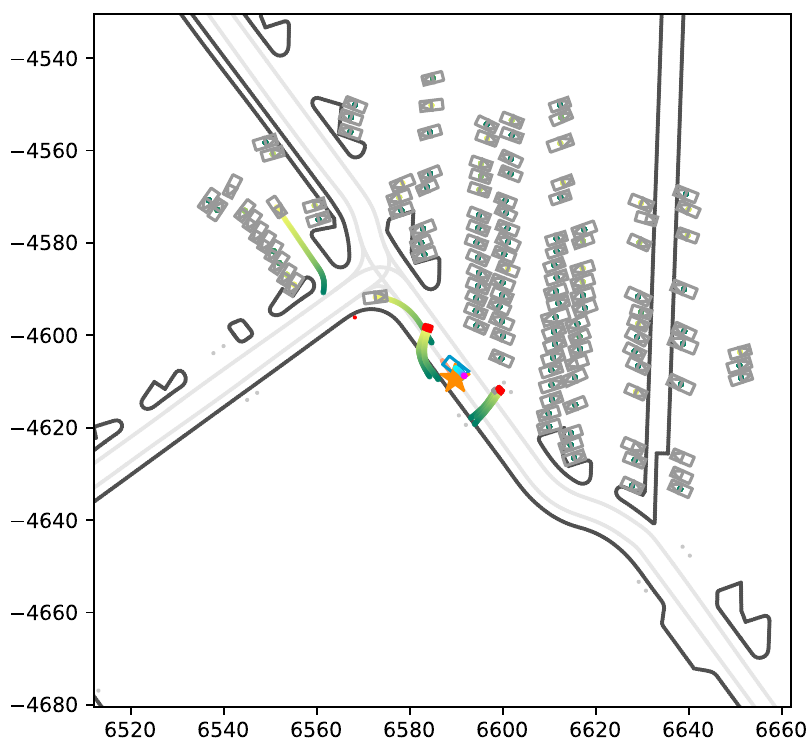}} \\
		\whline{0.5pt}\noalign{\smallskip} \multirow{2}{*}[+3em]{\addscentext{Curving Routing}} & {\addFig[trim=100 50 100 100, clip]{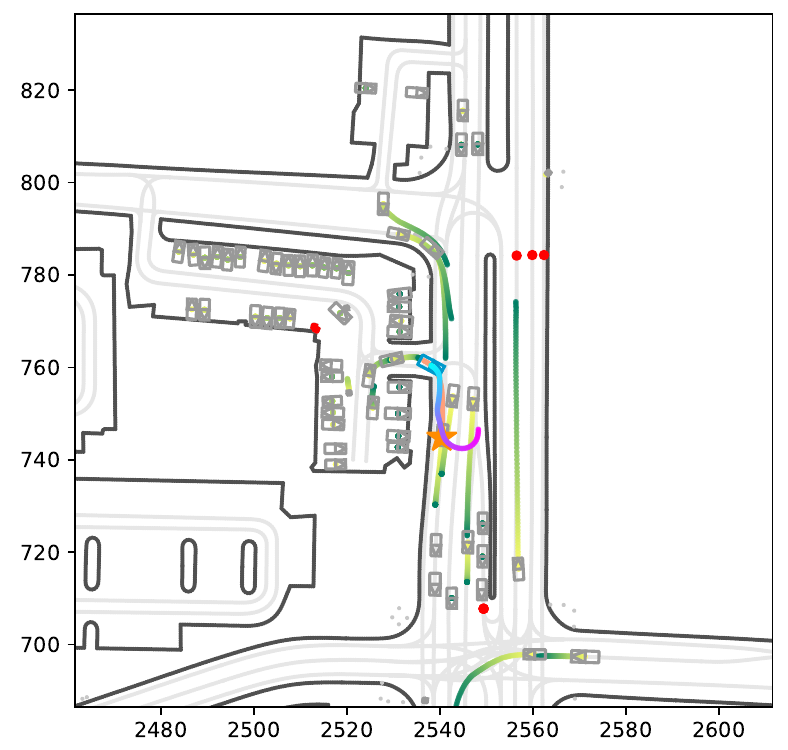}}              & {\addFig[trim=100 50 100 100, clip]{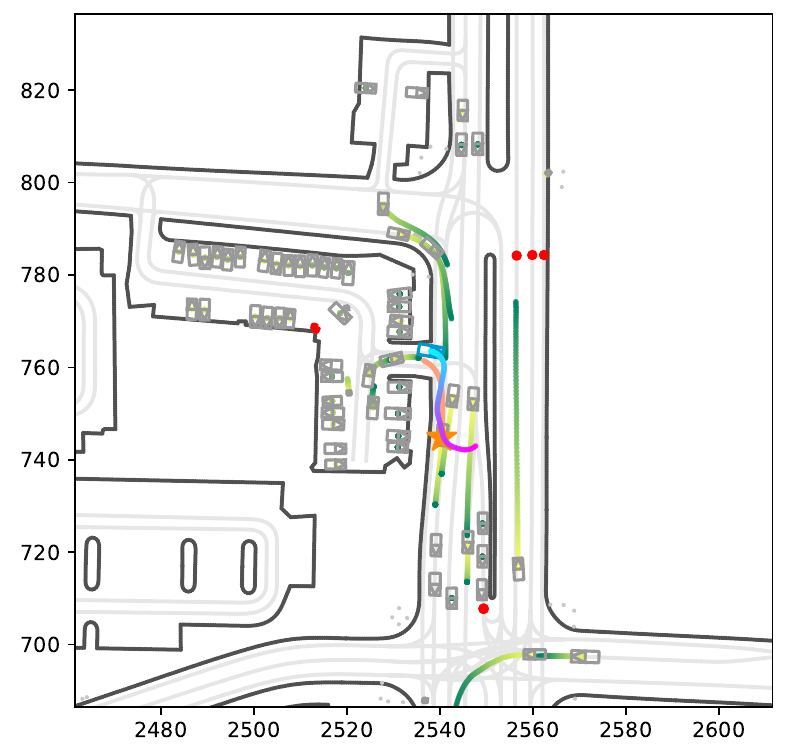}}  & {\addFig[trim=100 50 100 100, clip]{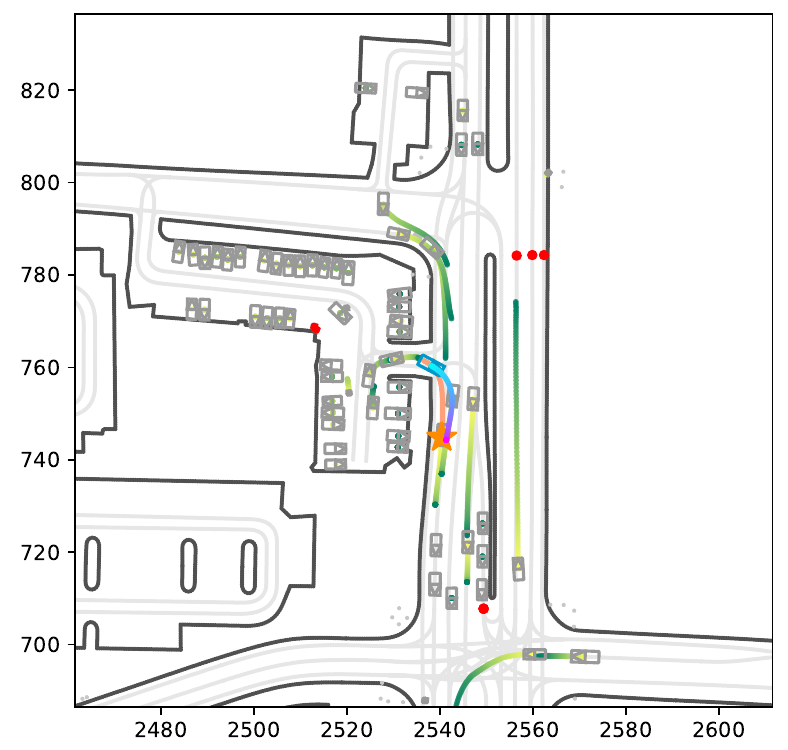}}  & {\addFig[trim=100 50 100 100, clip]{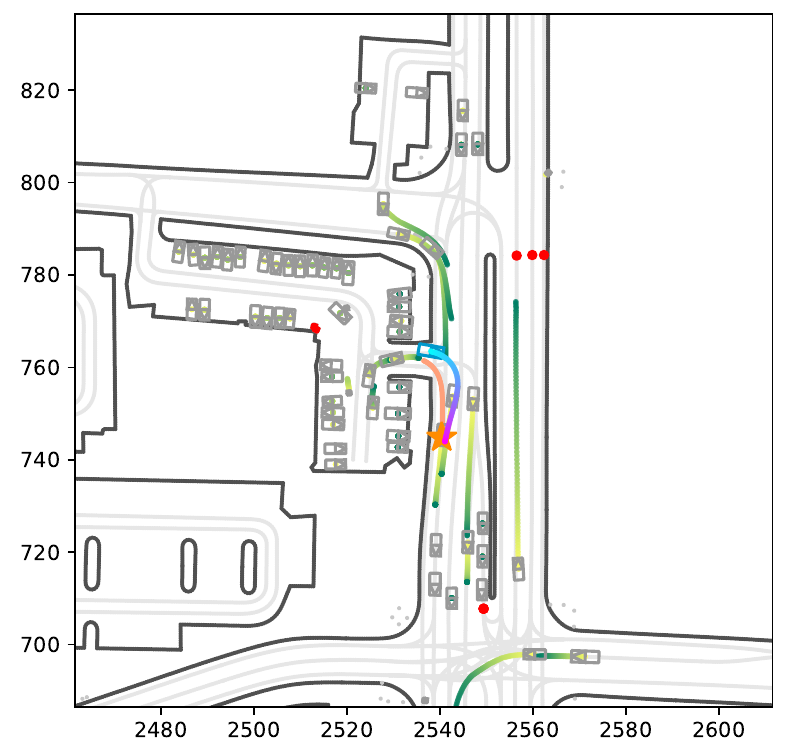}}  \\
		                                                                                        & {\addFig[trim=100 150 100 50, clip]{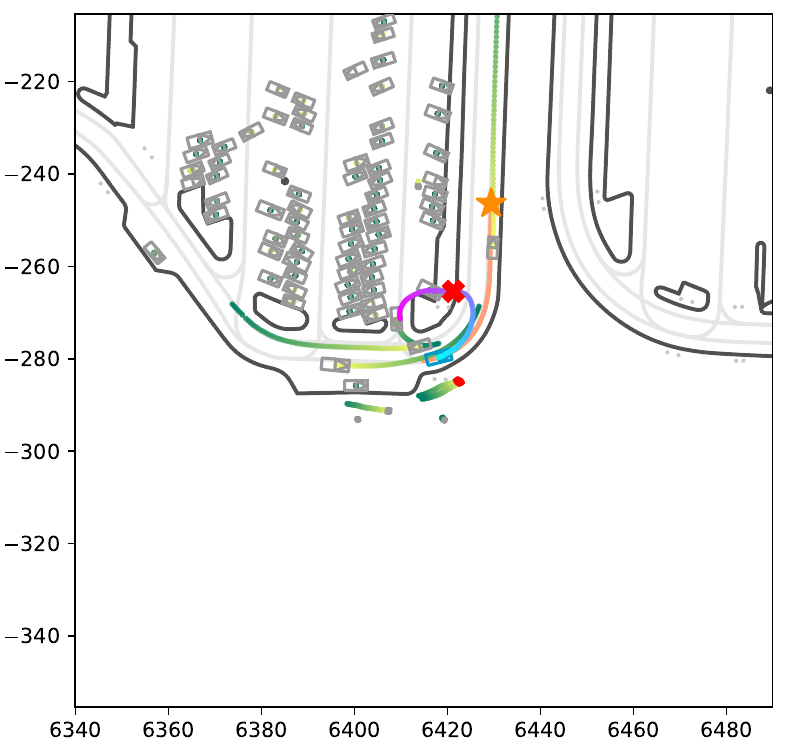}}              & {\addFig[trim=100 150 100 50, clip]{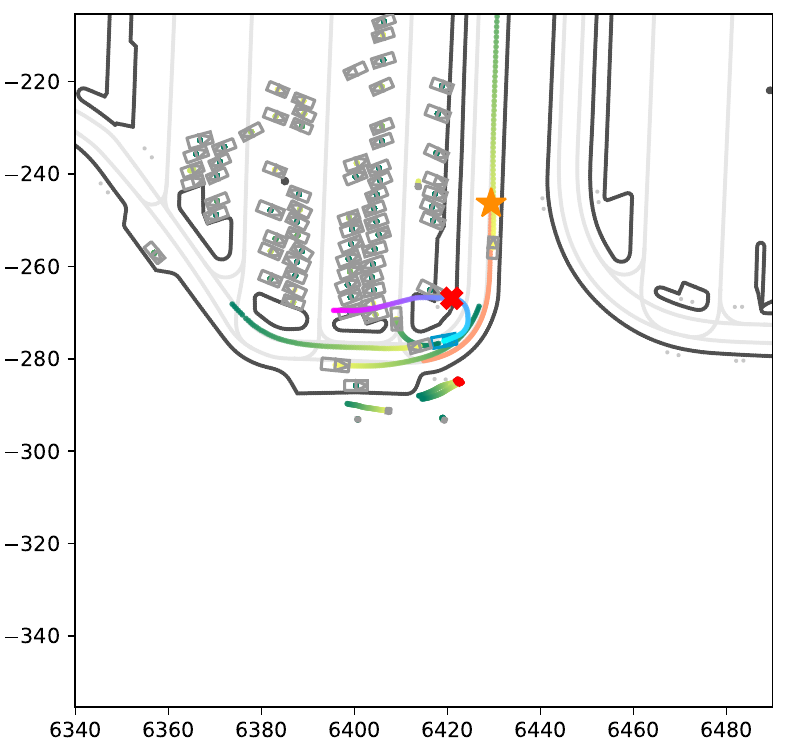}}  & {\addFig[trim=100 150 100 50, clip]{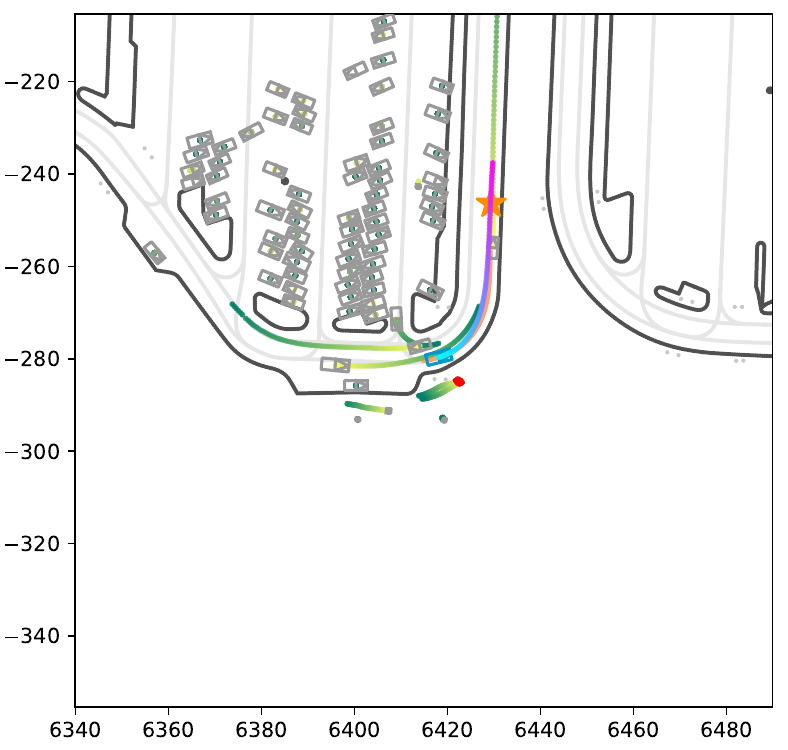}}  & {\addFig[trim=100 150 100 50, clip]{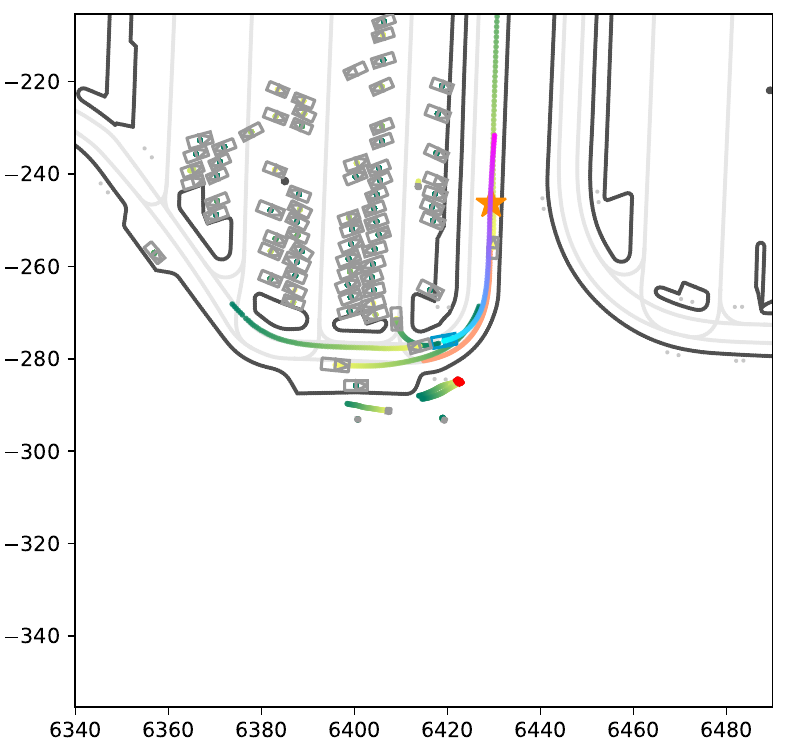}}  \\
		                                                                                        & \multicolumn{4}{c}{\includegraphics[width=\linewidth]{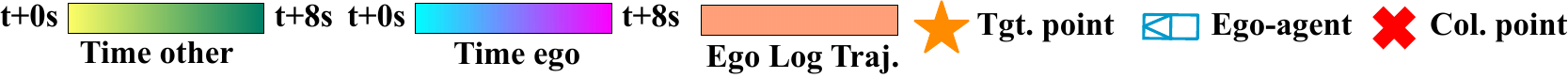}}
	\end{tabular}
	}
	\caption{Visualisation of \textit{close-loop} evaluation on parking lot. The
	initial scenario elucidates the process of navigating through a congested
	parking area. The subsequent scenario depicts the ego-agent advancing at a reduced
	pace, adhering to the requirement to yield to pedestrians. The third scenario illustrates
	the act of exiting the parking lot and integrating into a congested stream of traffic,
	ultimately halting due to traffic congestion. The final scenario demonstrates manoeuvring
	through a minor-angle turn within the confines of a narrow and congested parking
	area. }
	\label{fig.supp-lot}
\end{figure*}

\begin{figure*}[tp]
	\centering
	\resizebox{\linewidth}{!}{
	\begin{tabular}{ccccc}
		                                                      & \multicolumn{2}{c}{EasyChauffeur-IL}                   & \multicolumn{2}{c}{EasyChauffeur-PPO}                   \\
		                                                      & w/o ego-shifting                                       & w/ ego-shifting                                        & w/o ego-shifting                                       & w/ ego-shifting                                        \\
		\multirow{2}{*}[+3em]{\addscentext{Straight Routing}} & {\addFig[trim=450 50 400 550, clip]{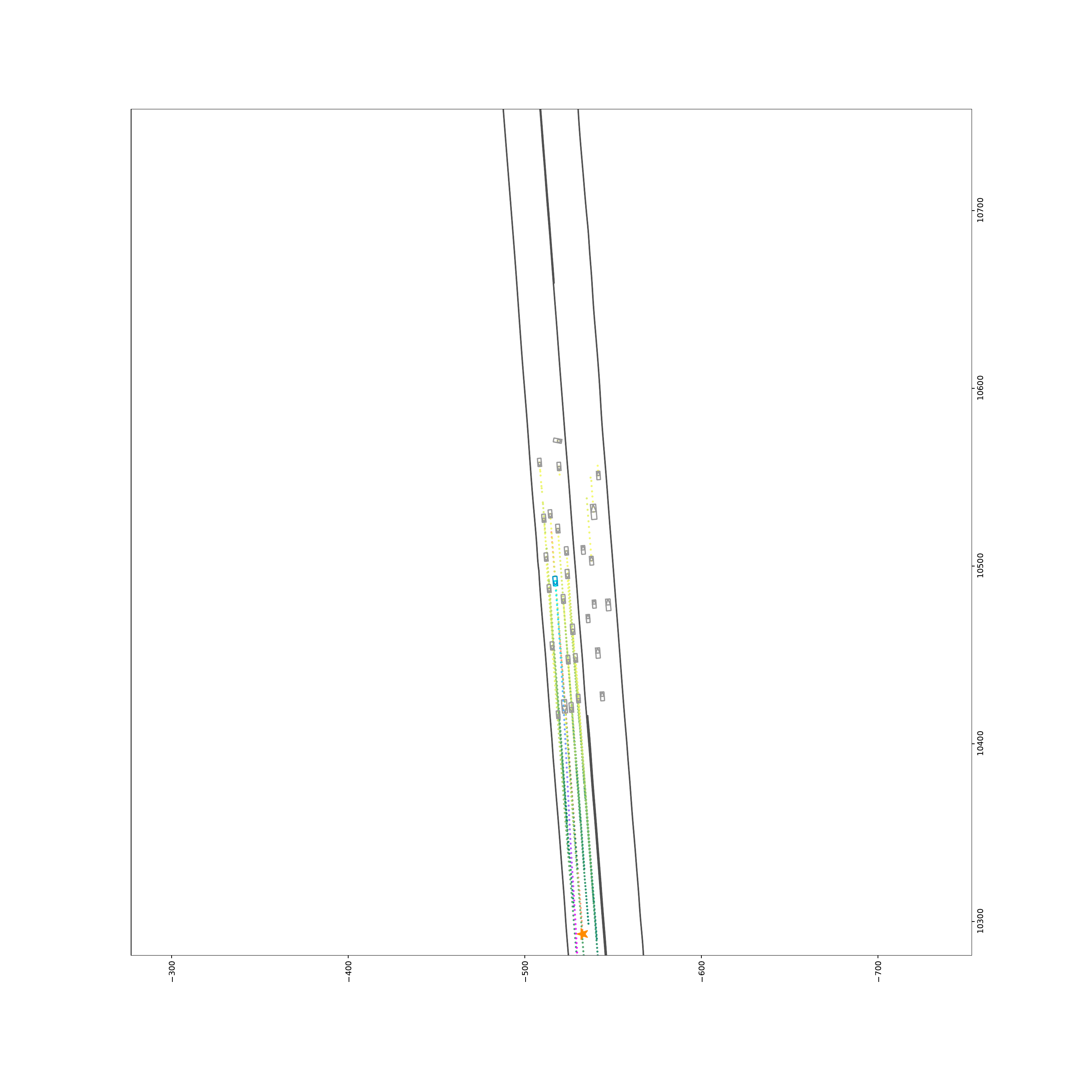}} & {\addFig[trim=450 50 400 550, clip]{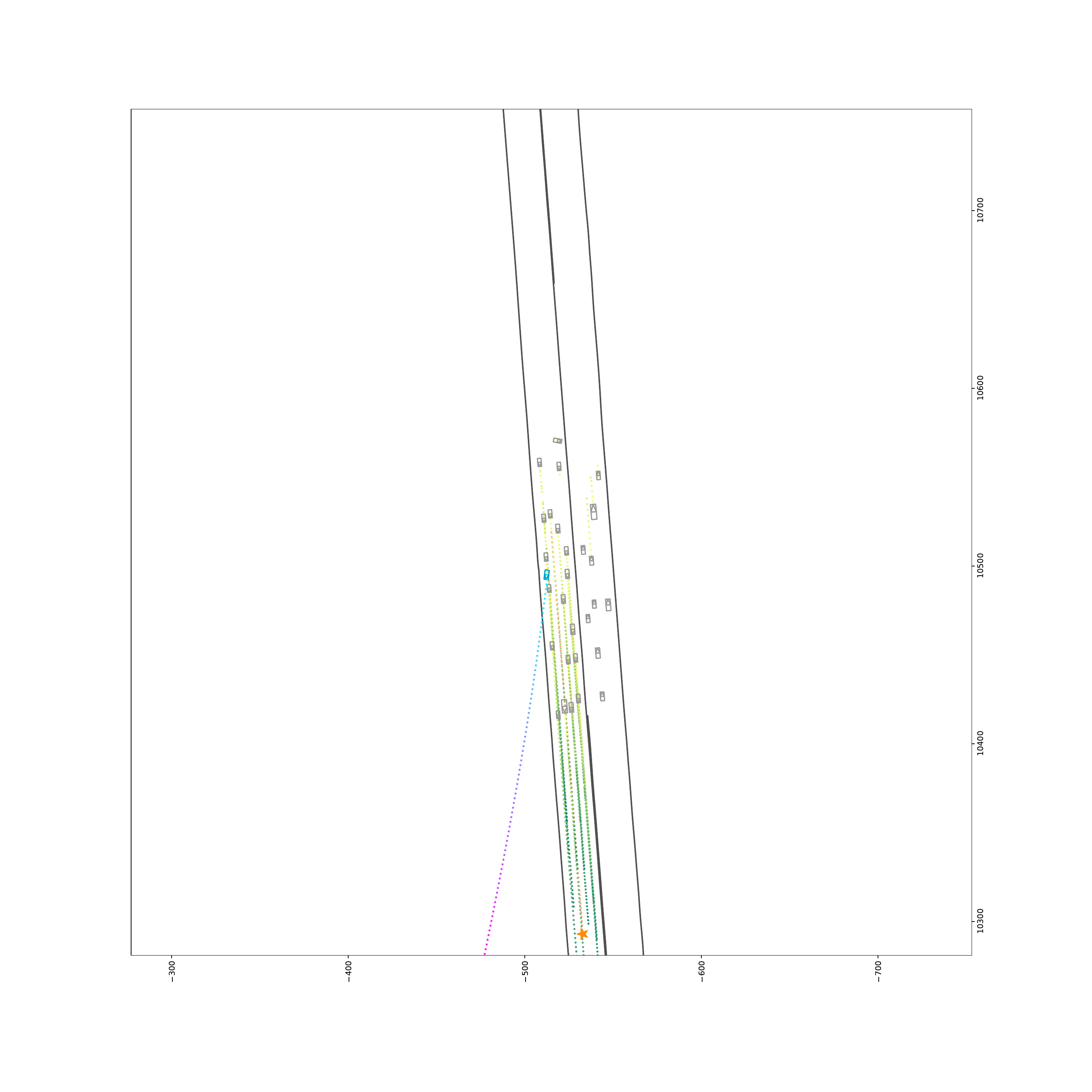}} & {\addFig[trim=450 60 400 550, clip]{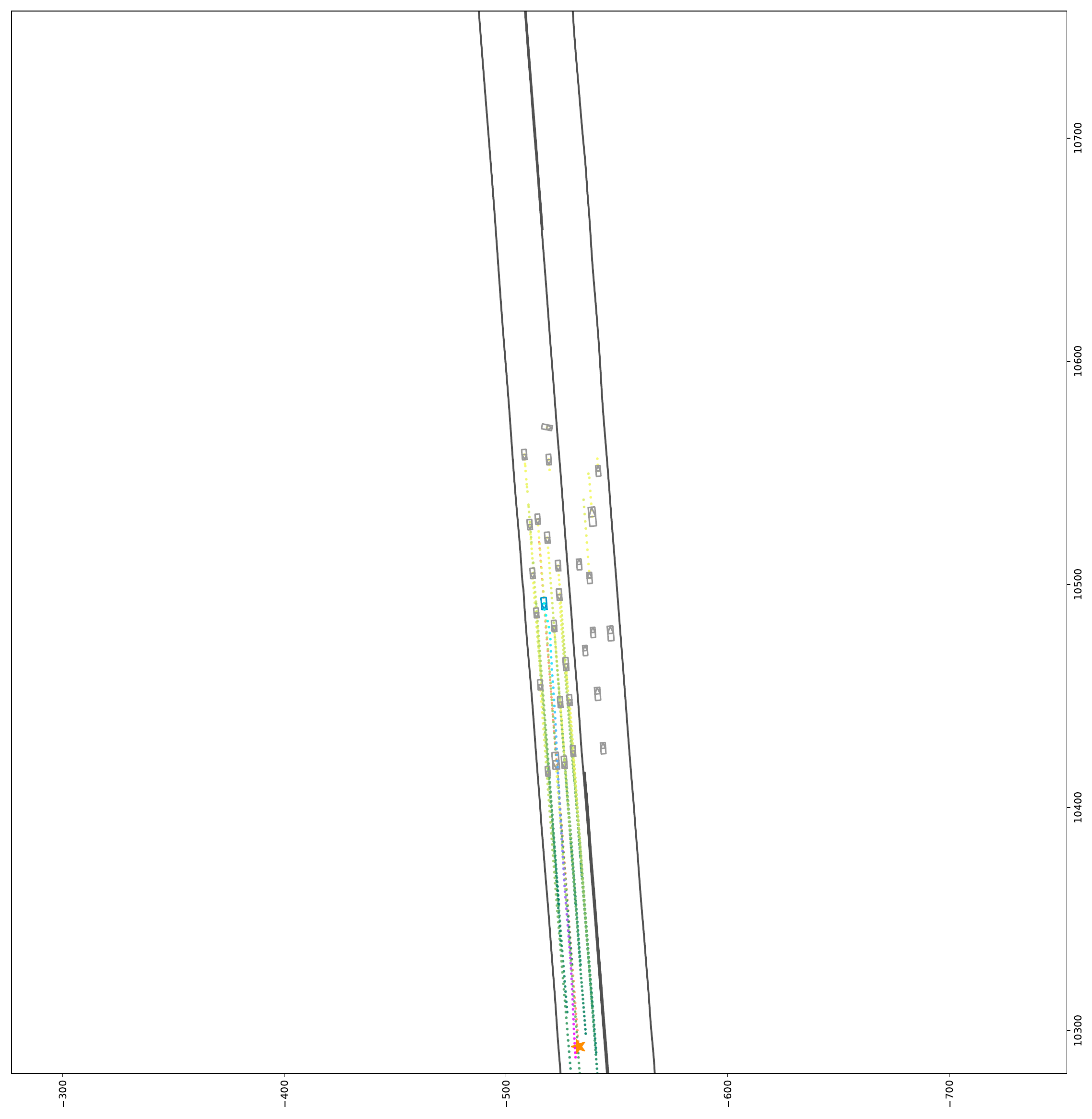}} & {\addFig[trim=450 50 400 550, clip]{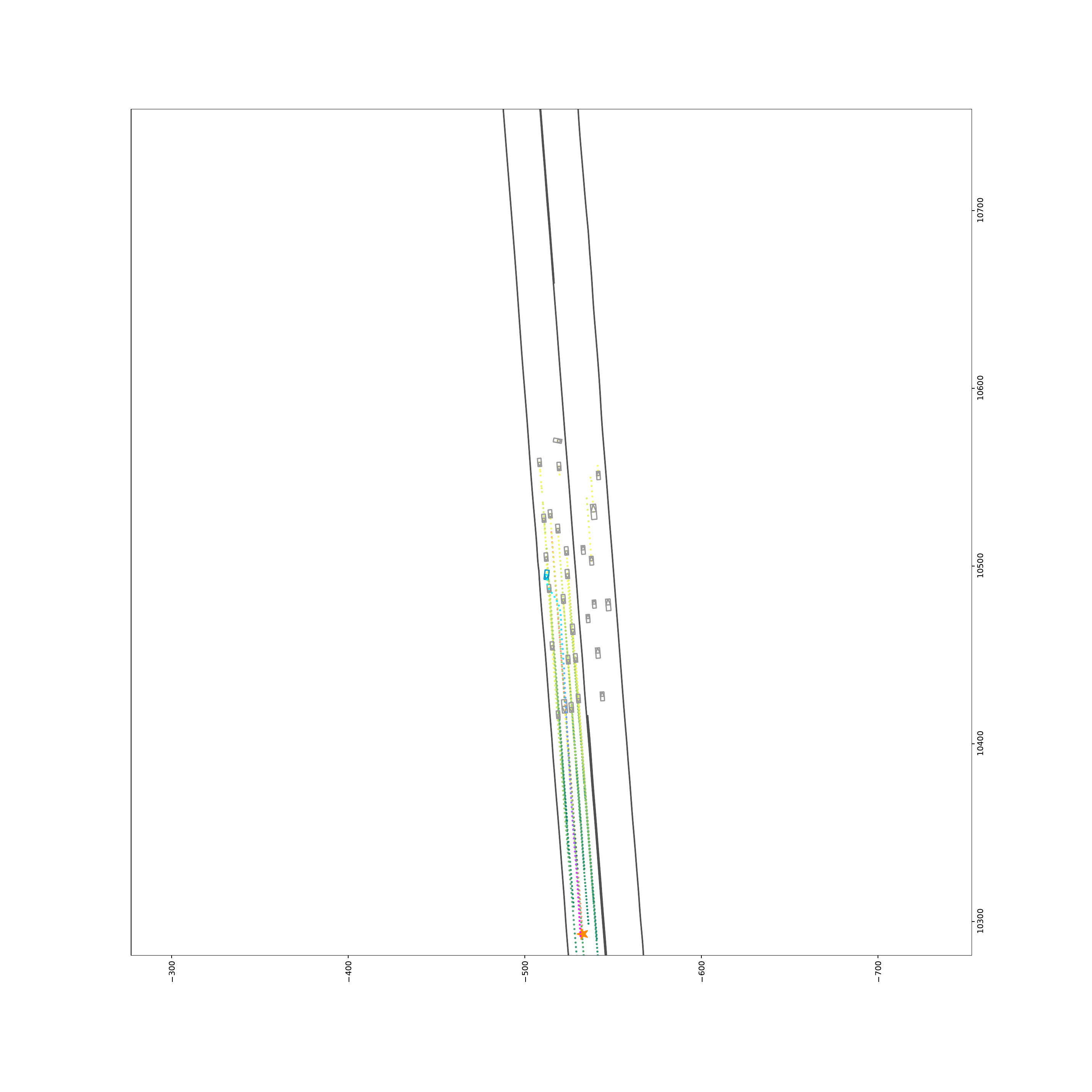}} \\
		                                                      & {\addFig[trim=270 10 170 200, clip]{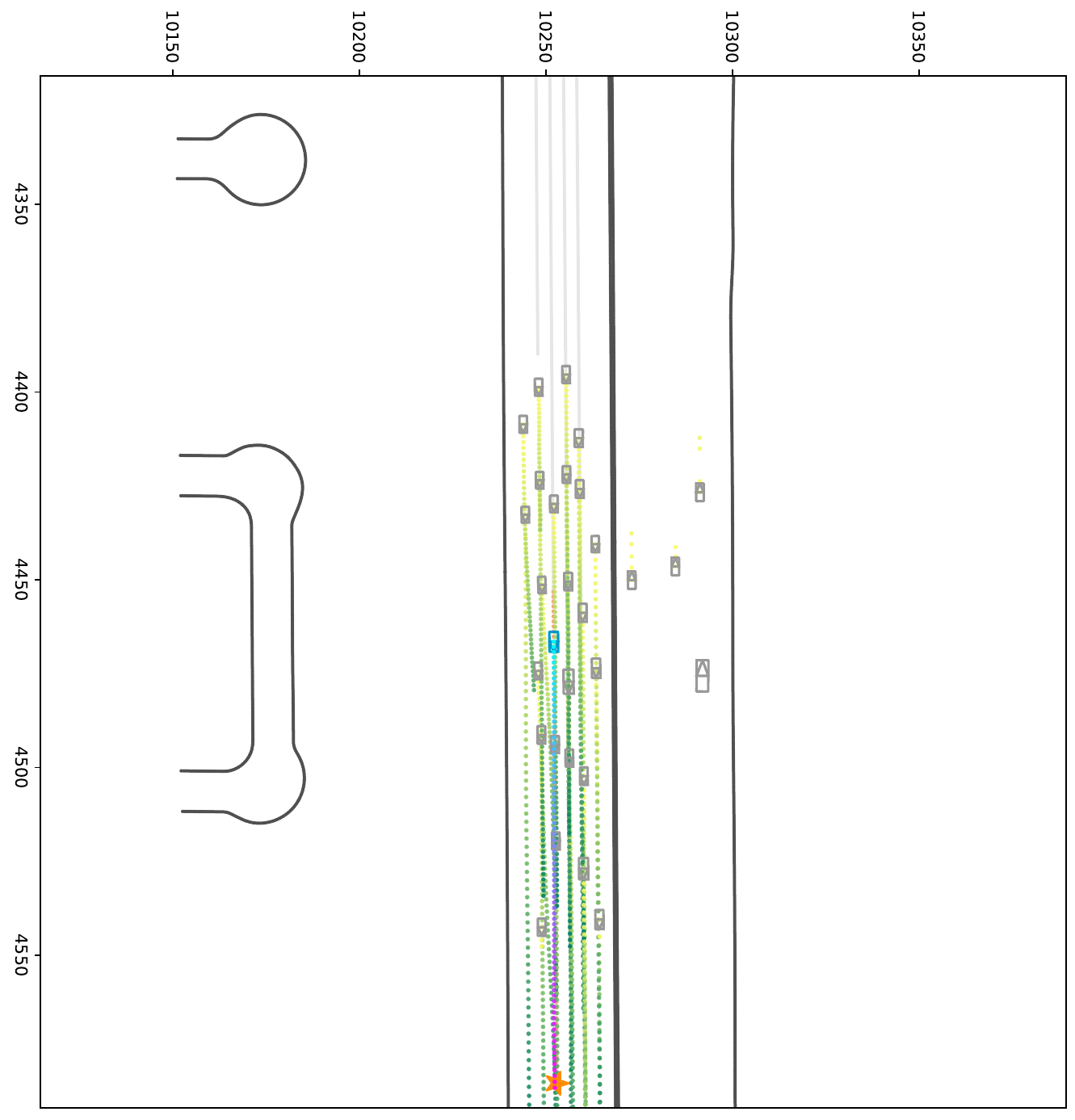}} & {\addFig[trim=270 10 170 200, clip]{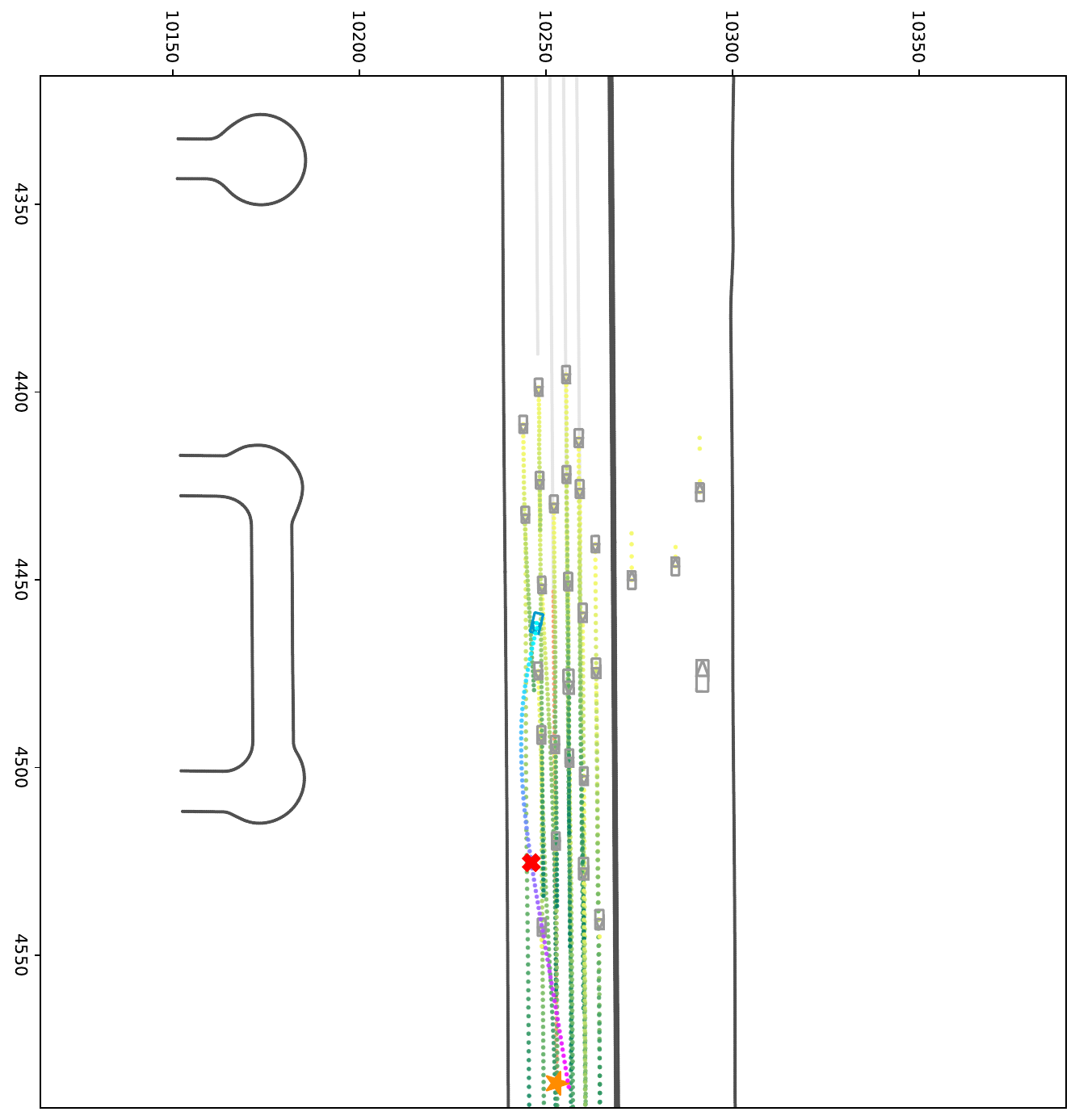}} & {\addFig[trim=270 10 170 200, clip]{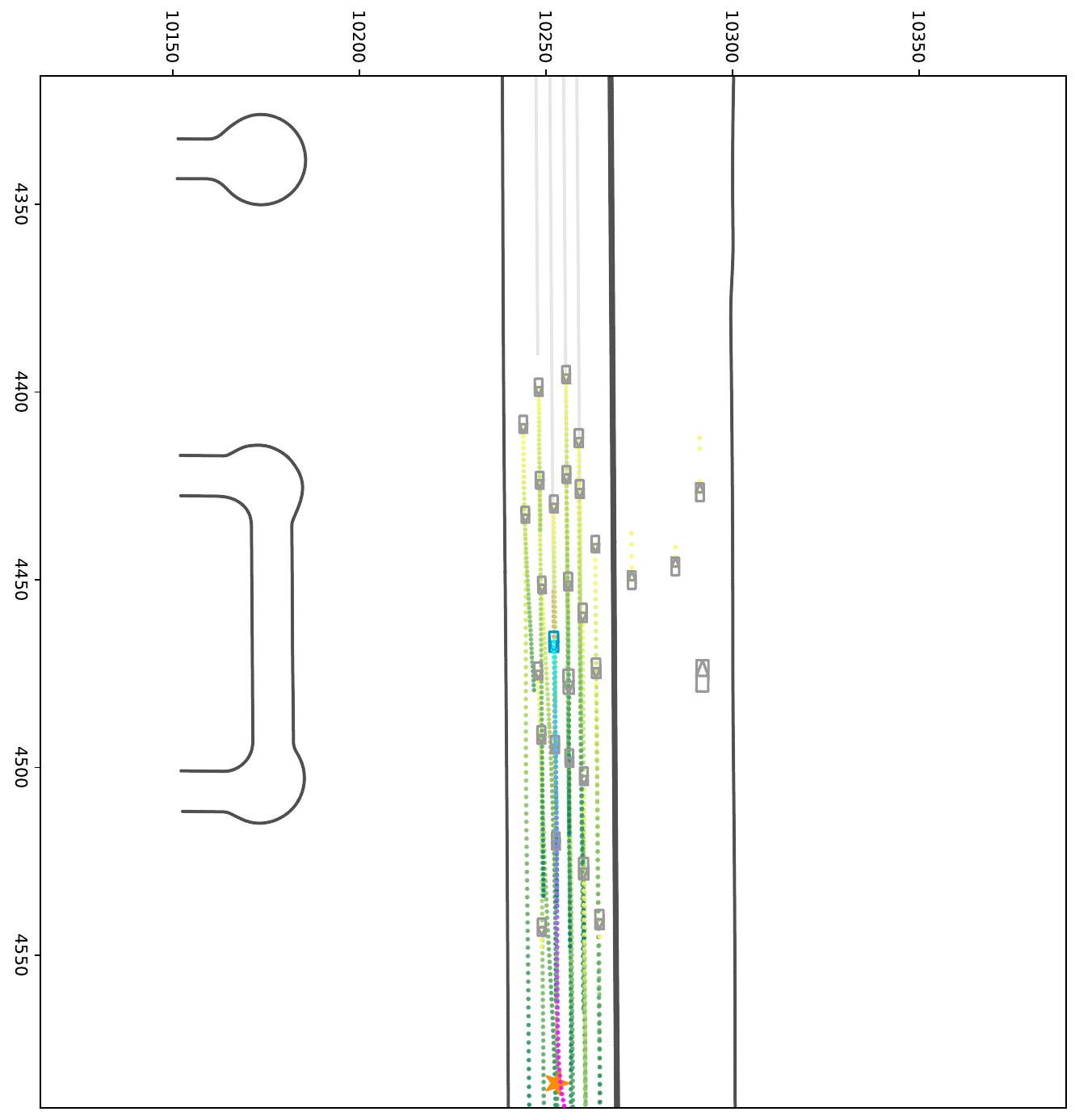}} & {\addFig[trim=270 10 170 200, clip]{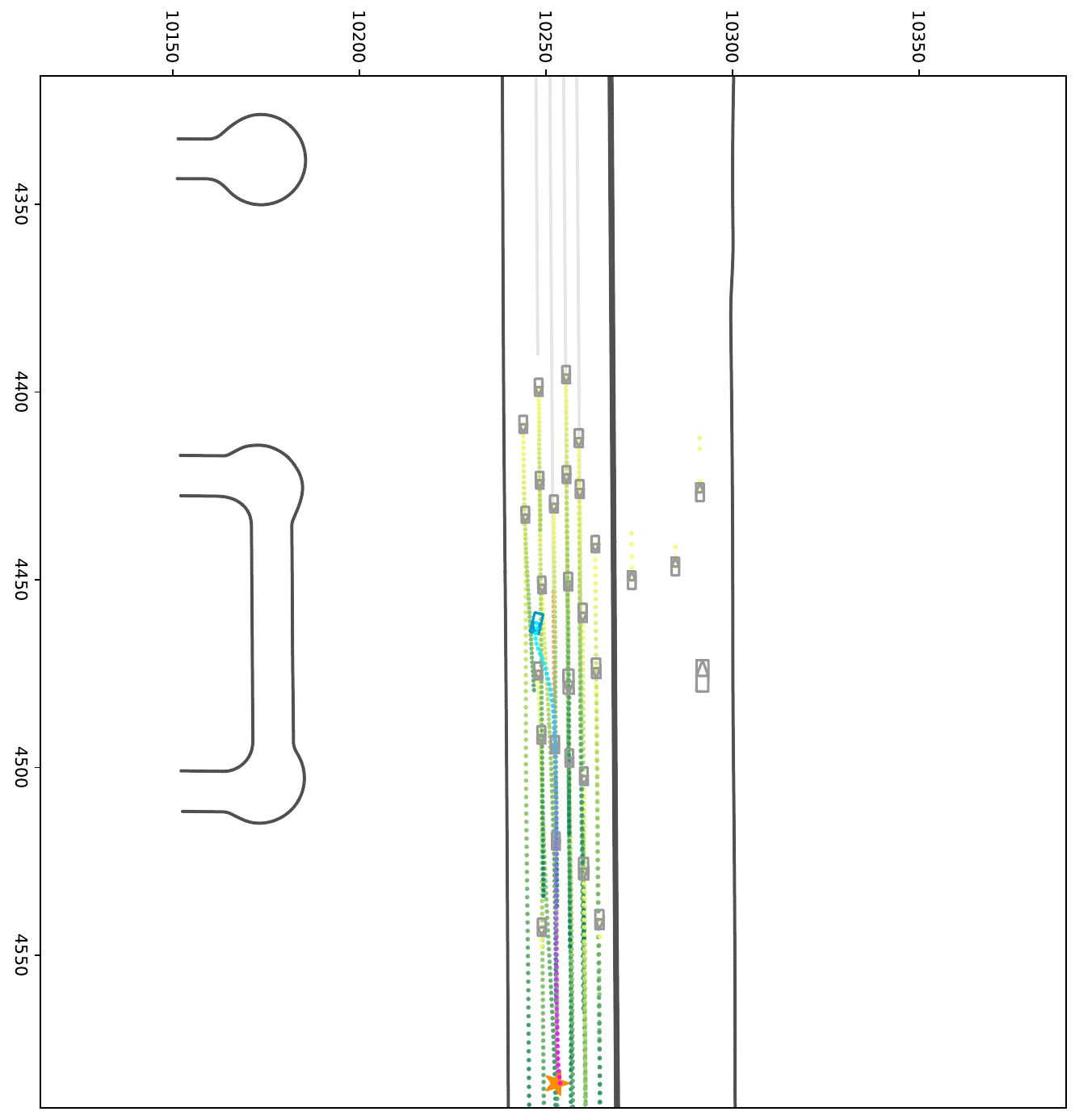}} \\
                                                              & \multicolumn{4}{c}{\includegraphics[width=\linewidth]{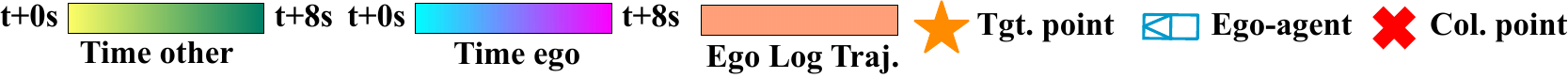}}
	\end{tabular}
	}
	\caption{Visualisation of \textit{close-loop} evaluation on dual carriageways.
	The two scenarios depict scenarios of high-speed traversal along
	complex, straight roads, characterised by the presence of numerous surrounding
	vehicles.}
	\label{fig.supp-highway-a}
\end{figure*}

\begin{figure*}[tp]
	\centering
	\resizebox{\linewidth}{!}{
	\begin{tabular}{ccccc}
		\multirow{2}{*}[+3em]{\addscentext{Curving Routing}} & {\addFig[trim=200 25 200 200, clip]{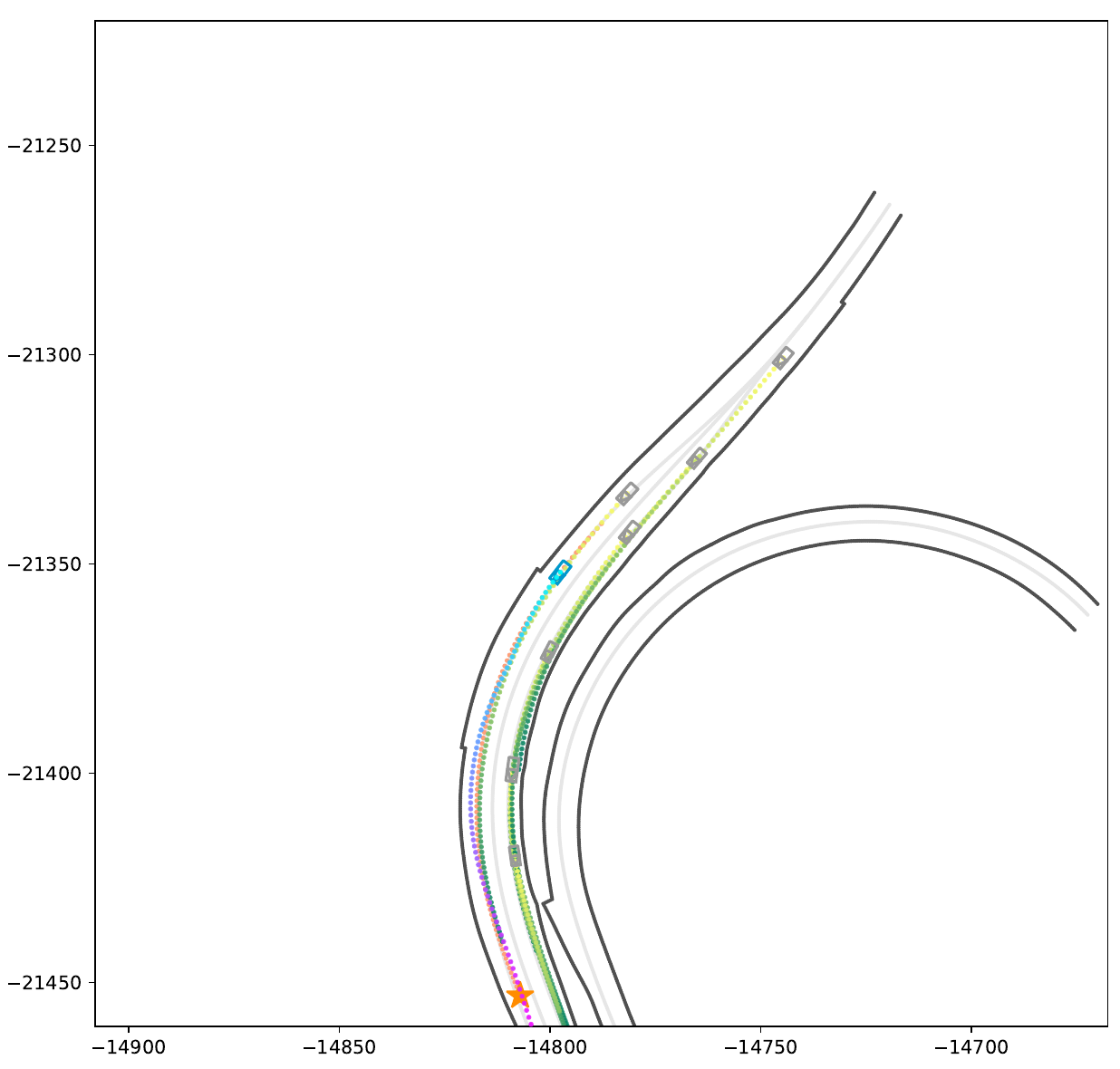}}                 & \addFig[trim=200 25 200 200, clip]{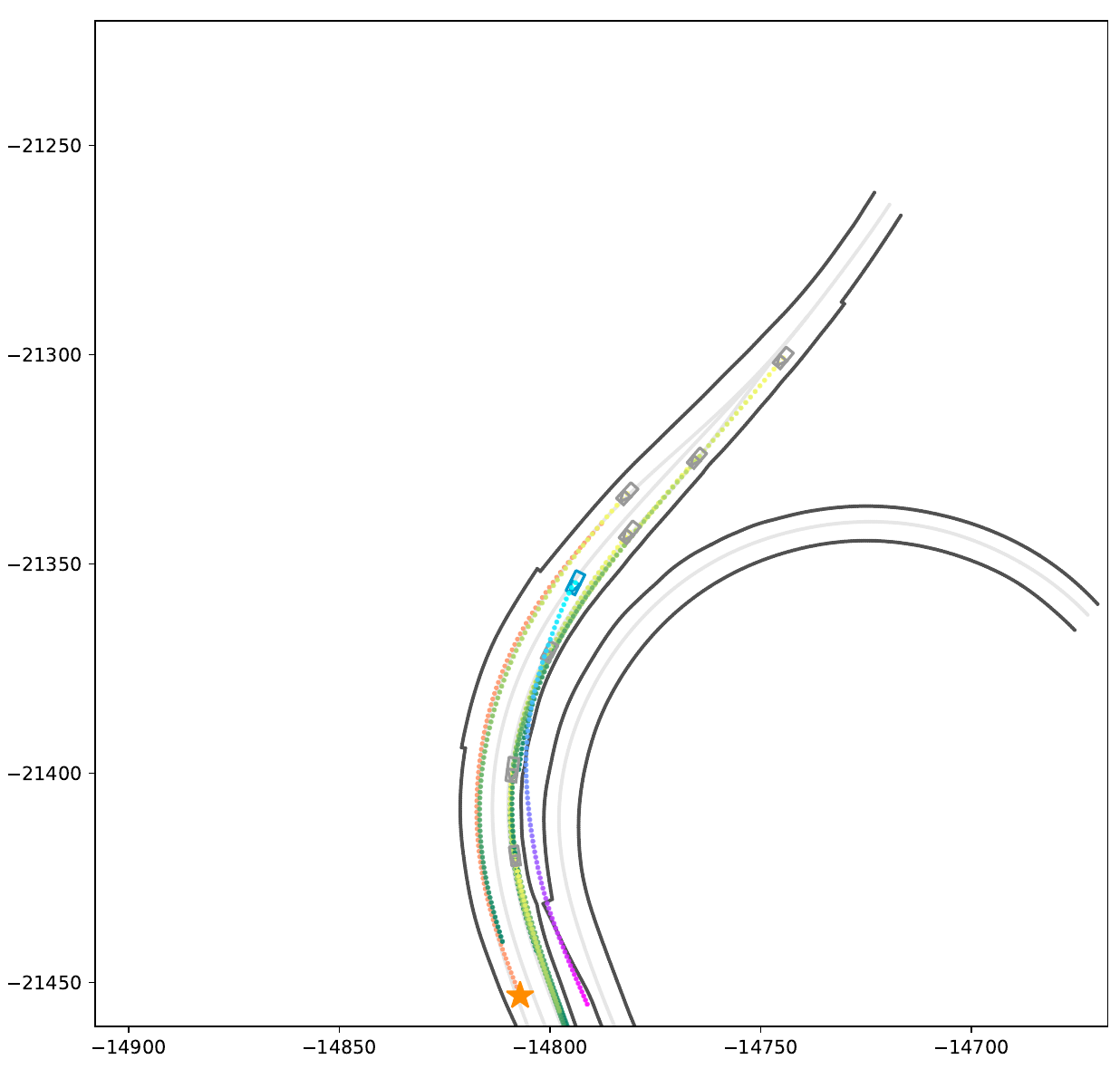} & \addFig[trim=200 25 200 200, clip]{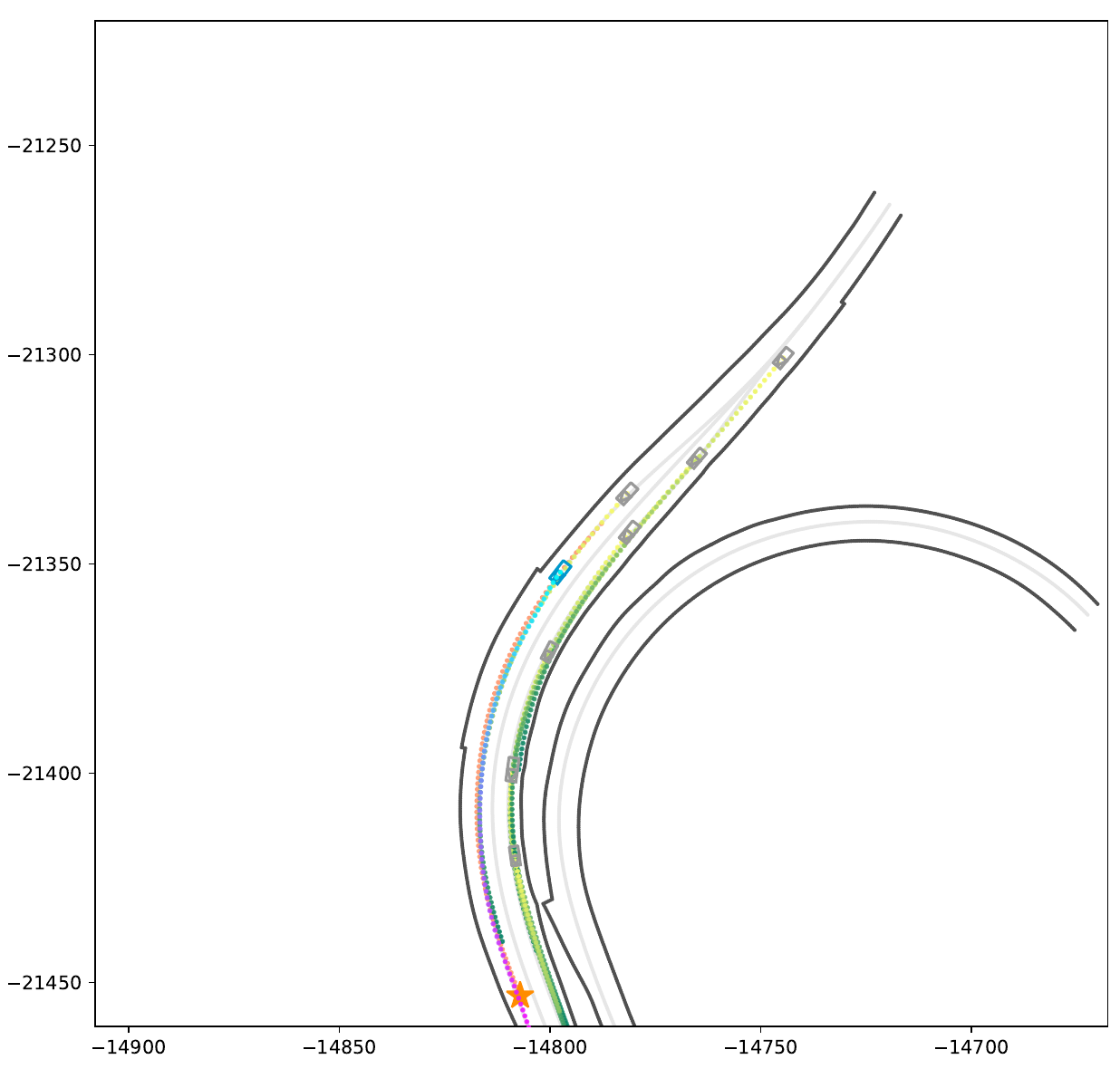} & \addFig[trim=200 34 200 200, clip]{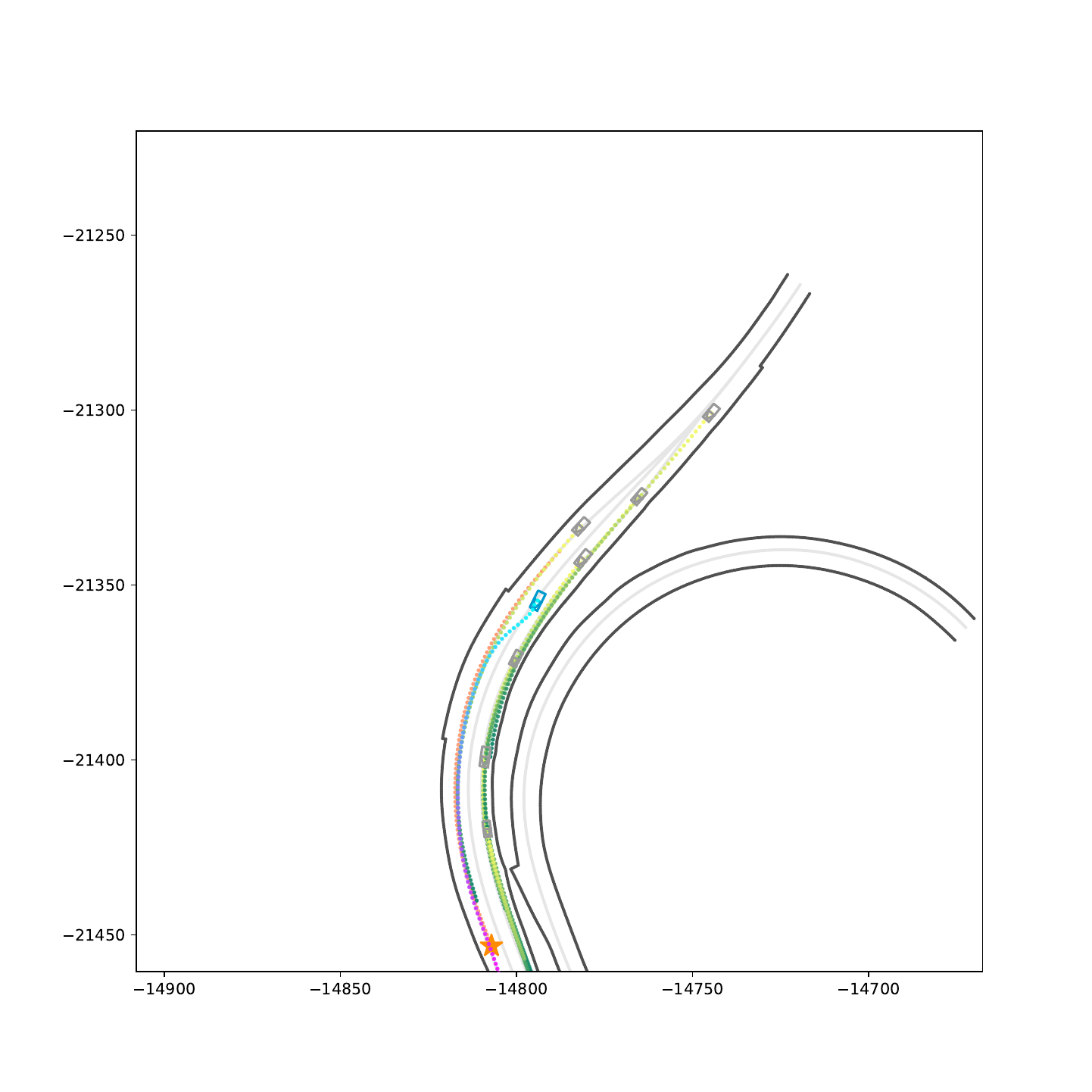} \\
		                                                     & \addFig[trim=50 50 200 50, clip]{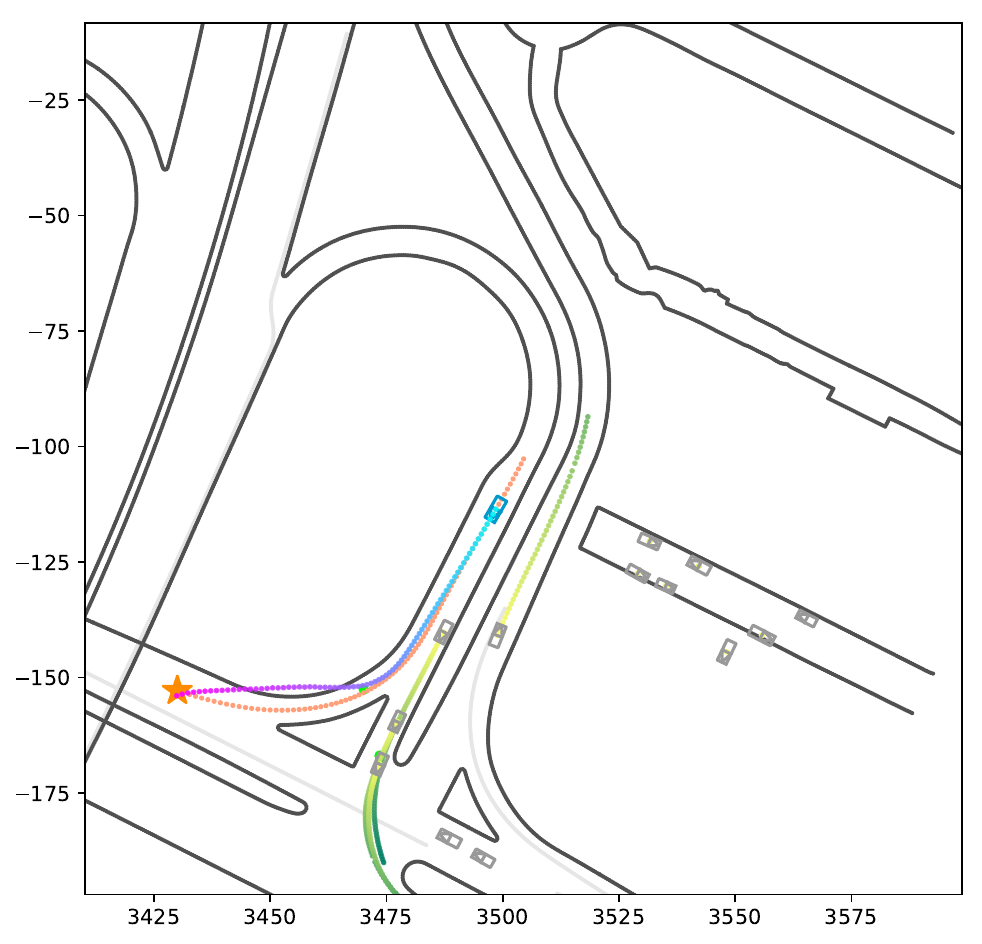}                     & \addFig[trim=50 50 200 50, clip]{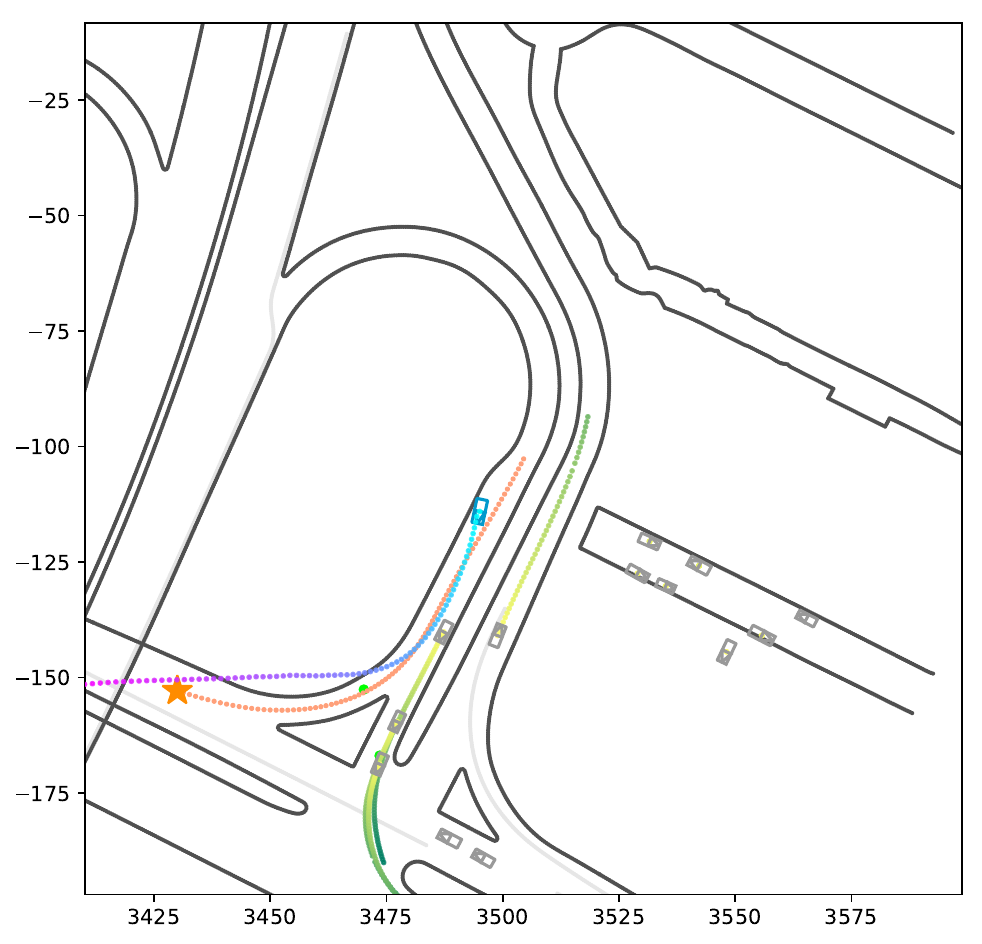}   & \addFig[trim=50 50 200 50, clip]{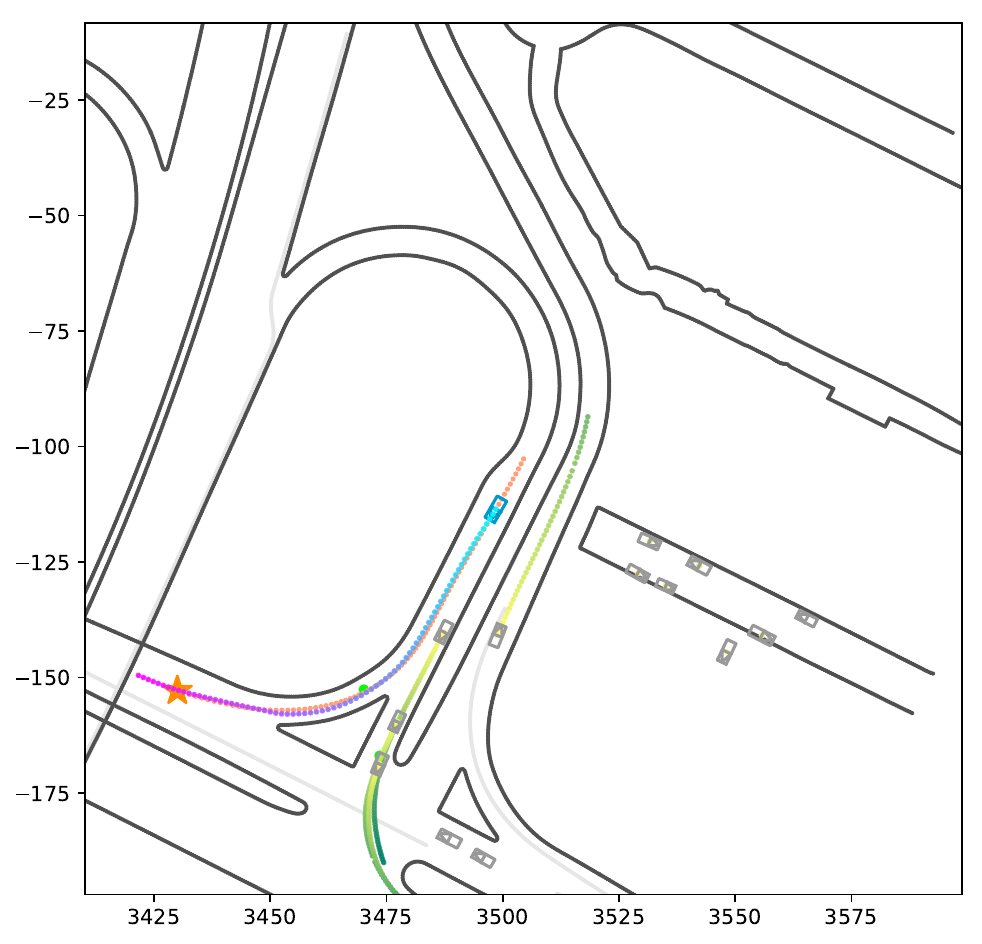}   & \addFig[trim=50 50 200 50, clip]{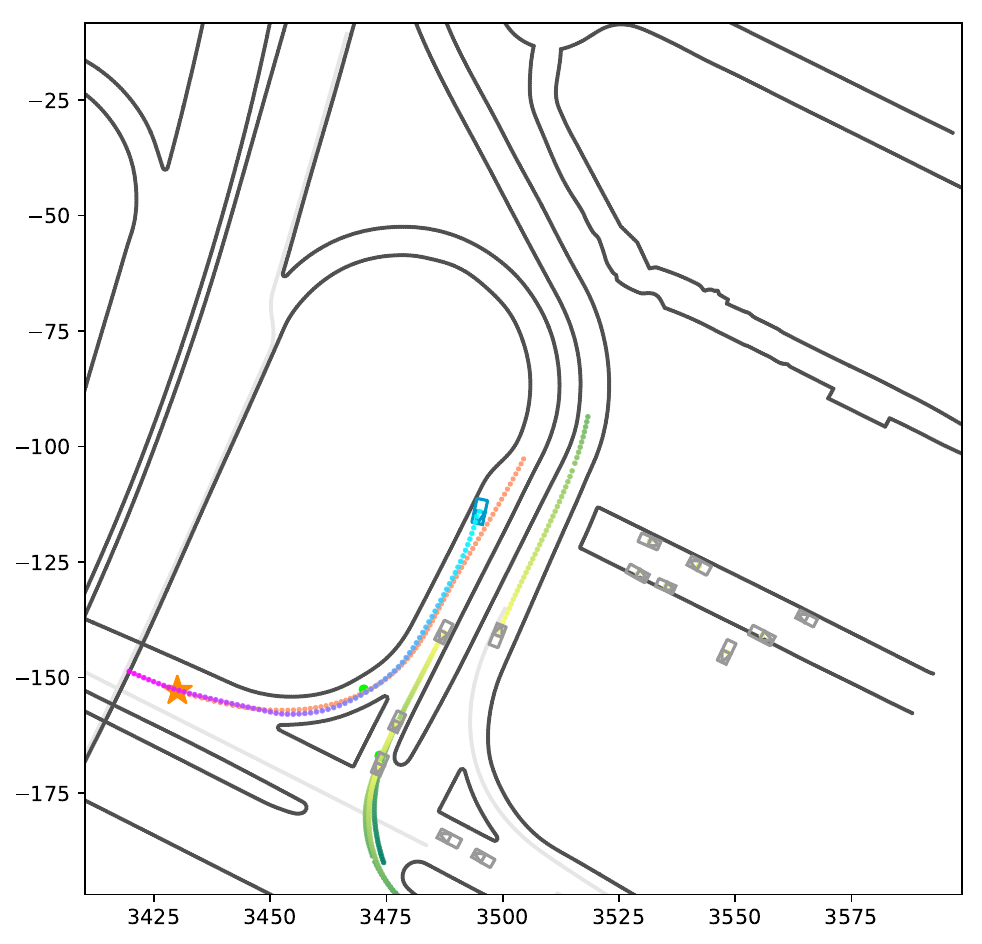}   \\
		                                                     & \multicolumn{4}{c}{\includegraphics[width=\linewidth]{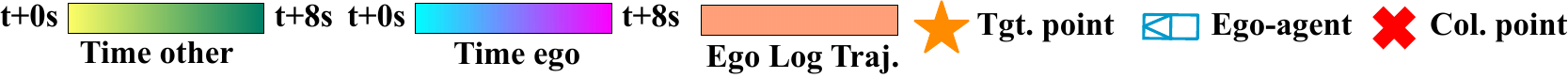}}
	\end{tabular}
	}
	\caption{Visualisation of \textit{close-loop} evaluation on dual carriageways.
	The imagery in the first row illustrates a section of
	motorways that is both curvaceous and narrow. The last row is dedicated to showcasing
	a slip road facilitating exit the dual carriageways. Notably, from the depiction
	in the two examples, there is an observable increase in curvature. }
	\label{fig.supp-highway-b}
\end{figure*}

\begin{figure*}[tp]
	\centering
	\resizebox{\linewidth}{!}{
	\begin{tabular}{ccccc}
		                                                                                         & \multicolumn{2}{c}{EasyChauffeur-IL}                                   & \multicolumn{2}{c}{EasyChauffeur-PPO}                 \\
		                                                                                         & w/o ego-shifting                                                       & w/ ego-shifting                                      & w/o ego-shifting                                     & w/ ego-shifting                                      \\
		\multirow{2}{*}[+3em]{\addscentext{Straight Routing}}                                    & \addFig[trim=100 50 100 100, clip]{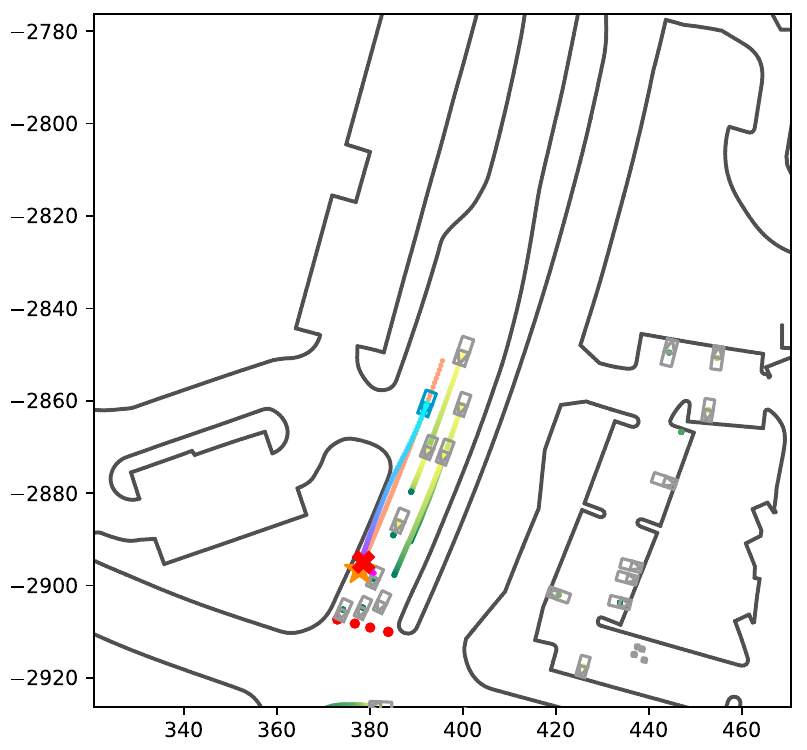}                   & \addFig[trim=100 50 100 100, clip]{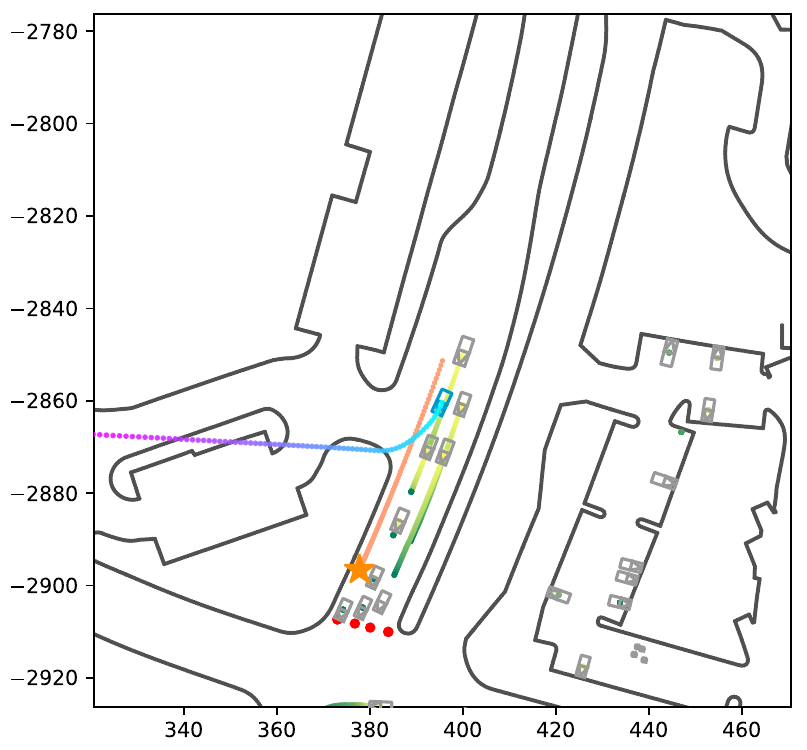} & \addFig[trim=100 50 100 100, clip]{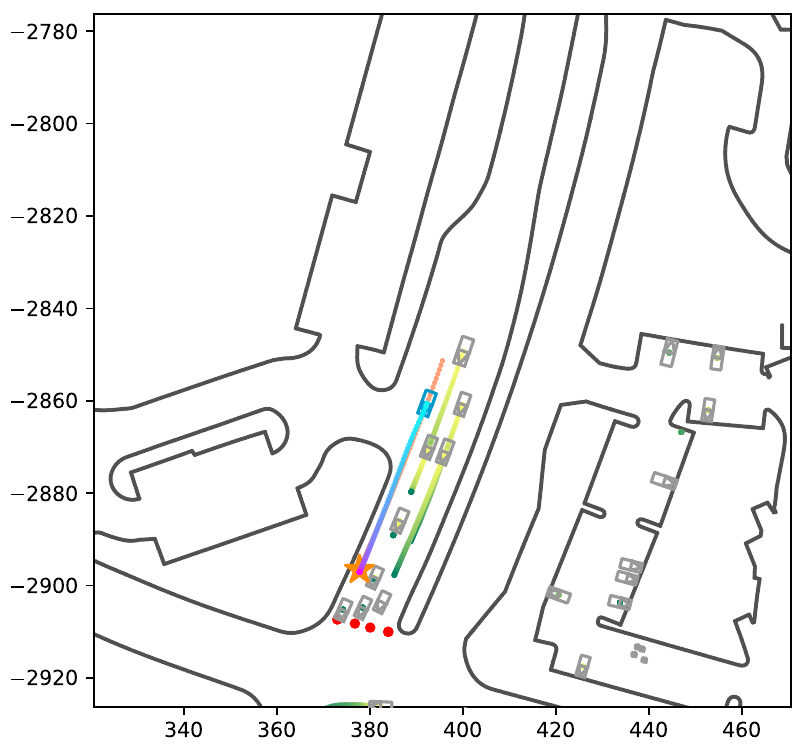} & \addFig[trim=100 50 100 100, clip]{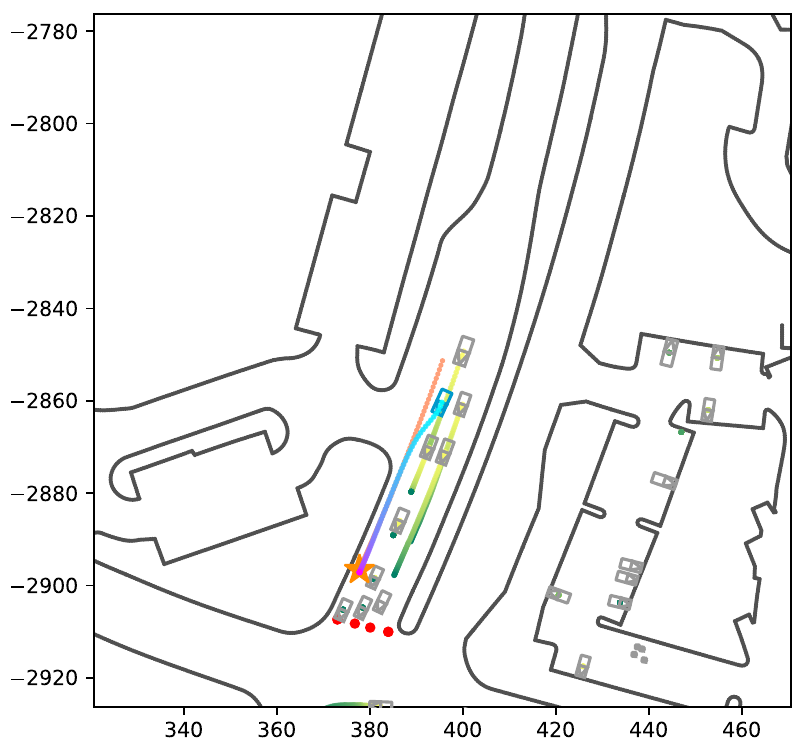} \\
		                                                                                         & \addFig[trim=100 100 20 100, clip]{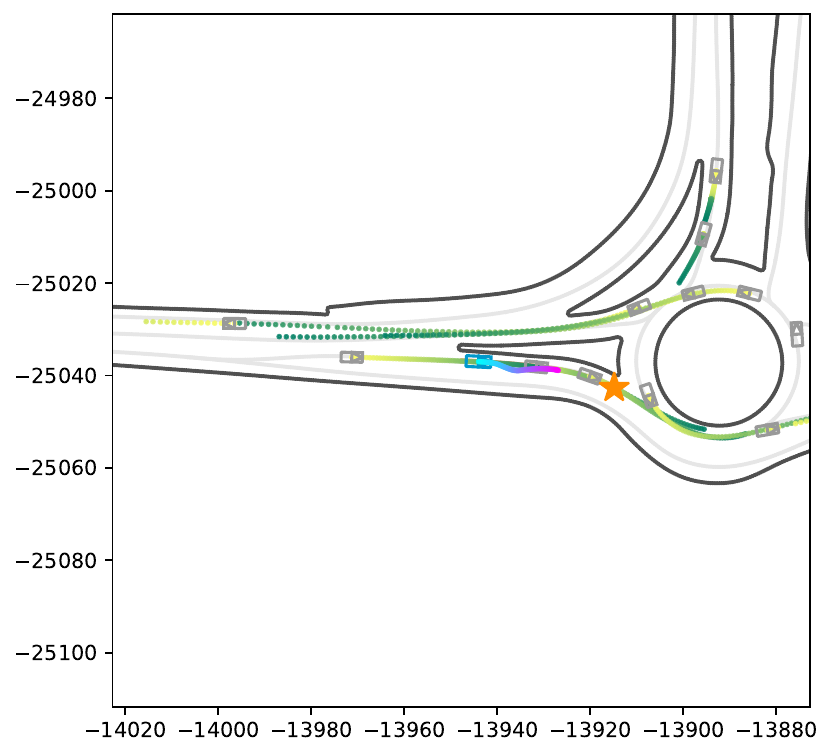}                     & \addFig[trim=100 100 20 100, clip]{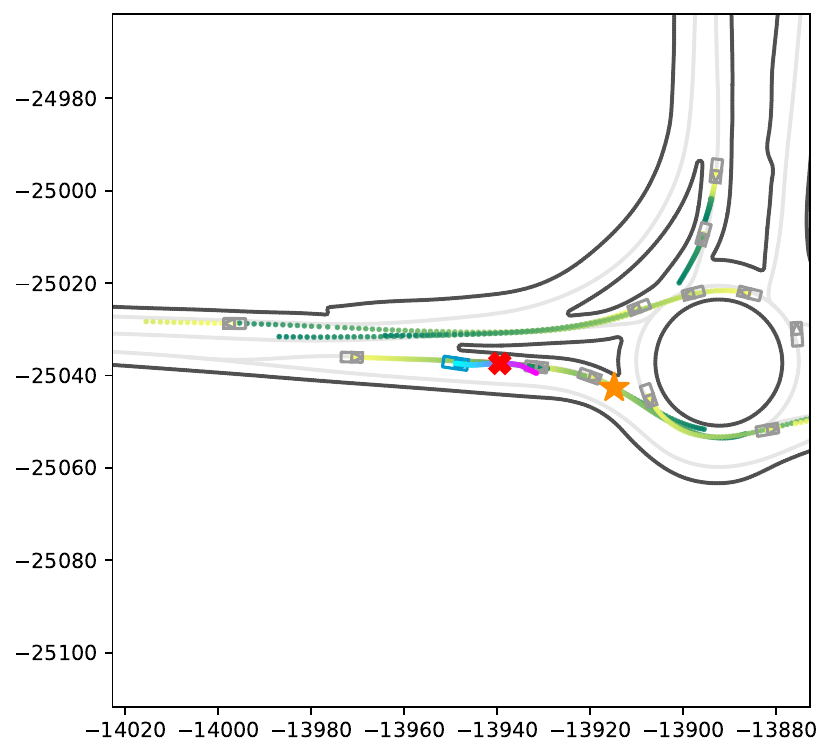}   & \addFig[trim=100 100 20 100, clip]{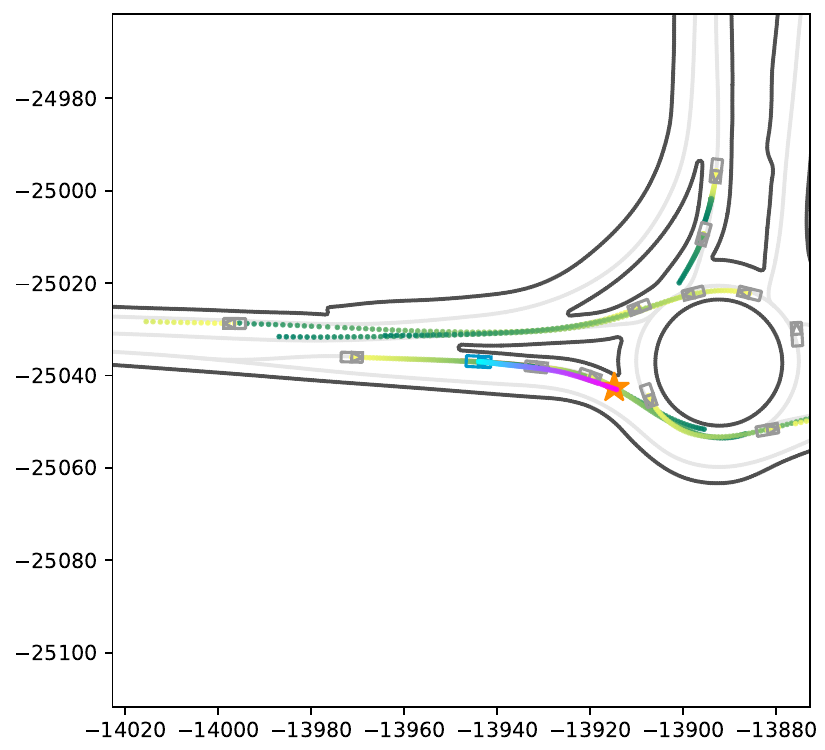}   & \addFig[trim=100 100 20 100, clip]{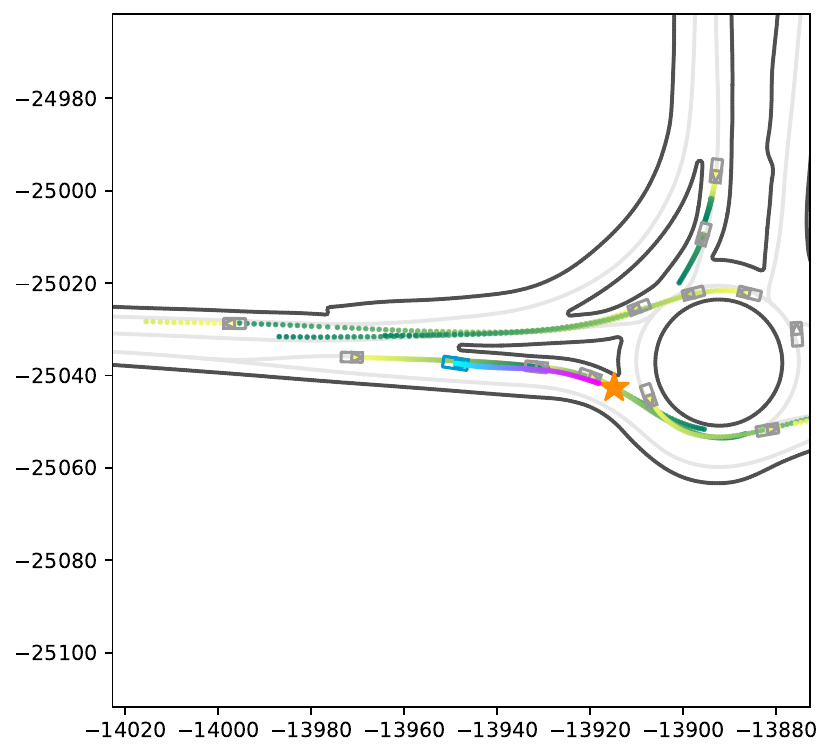}   \\
		\whline{0.5pt} \noalign{\smallskip} \multirow{2}{*}[+3em]{\addscentext{Curving Routing}} & \addFig[trim=150 150 50 20, clip]{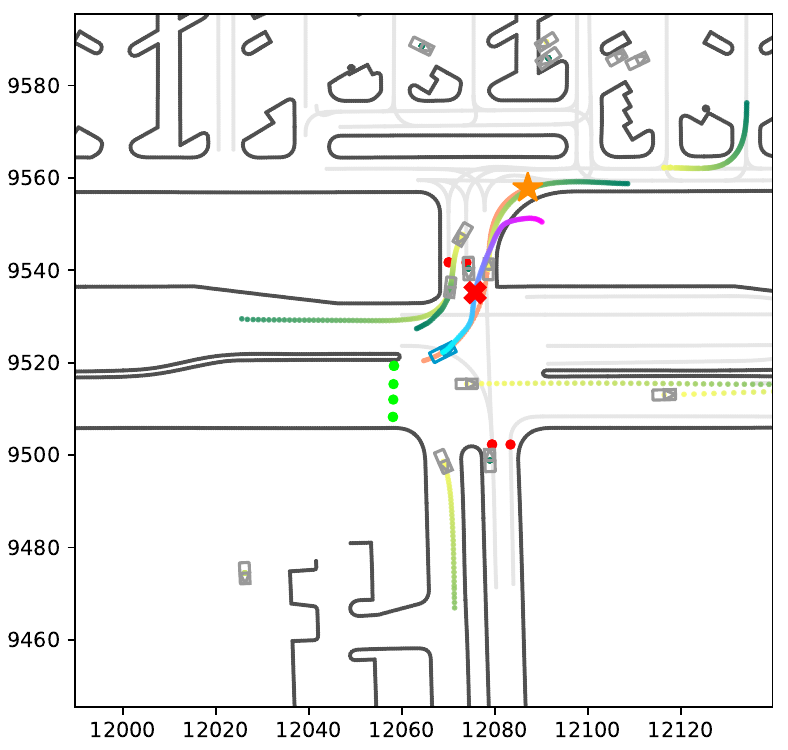}                      & \addFig[trim=150 150 50 20, clip]{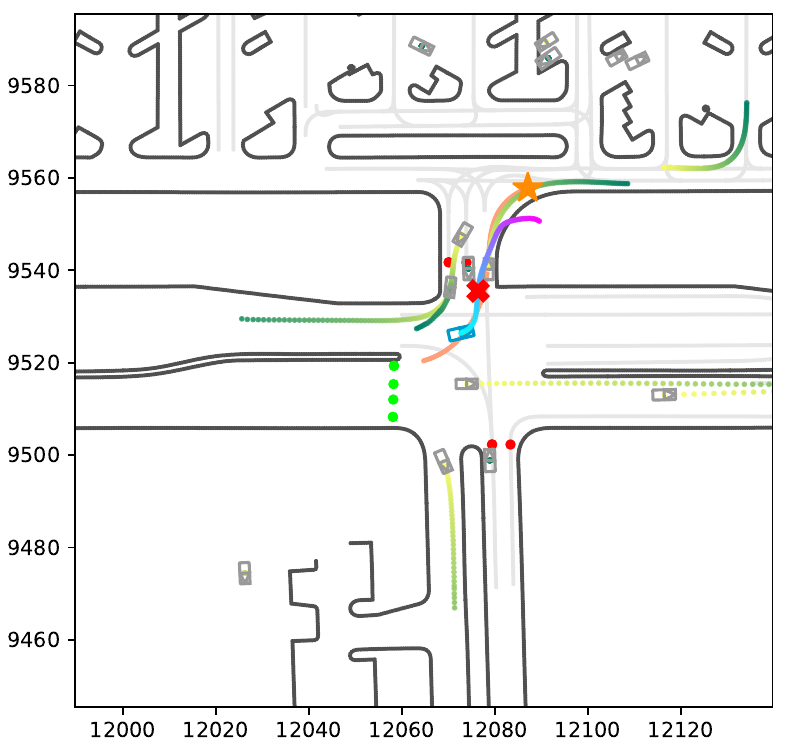}    & \addFig[trim=150 150 50 20, clip]{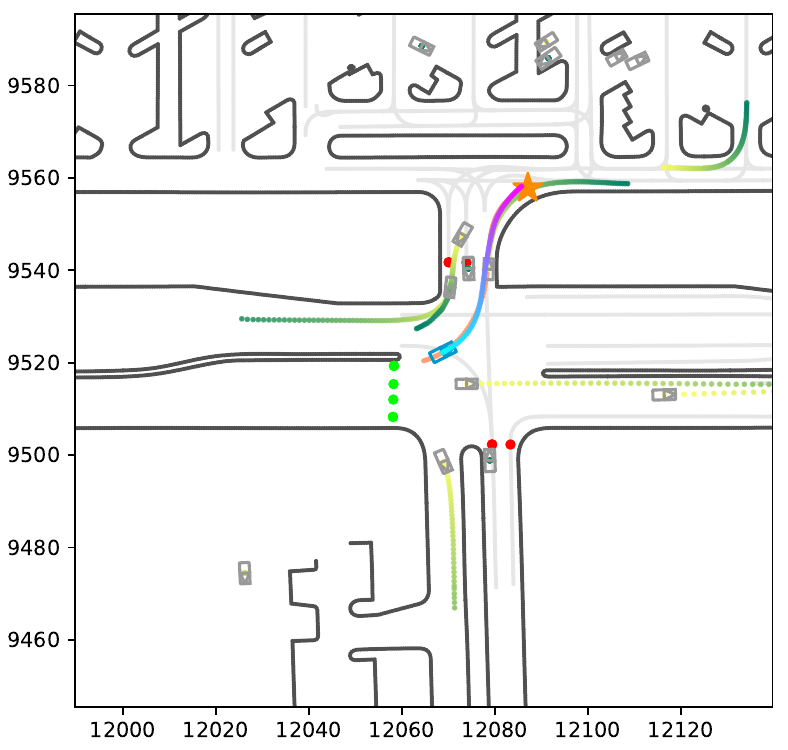}    & \addFig[trim=150 150 50 20, clip]{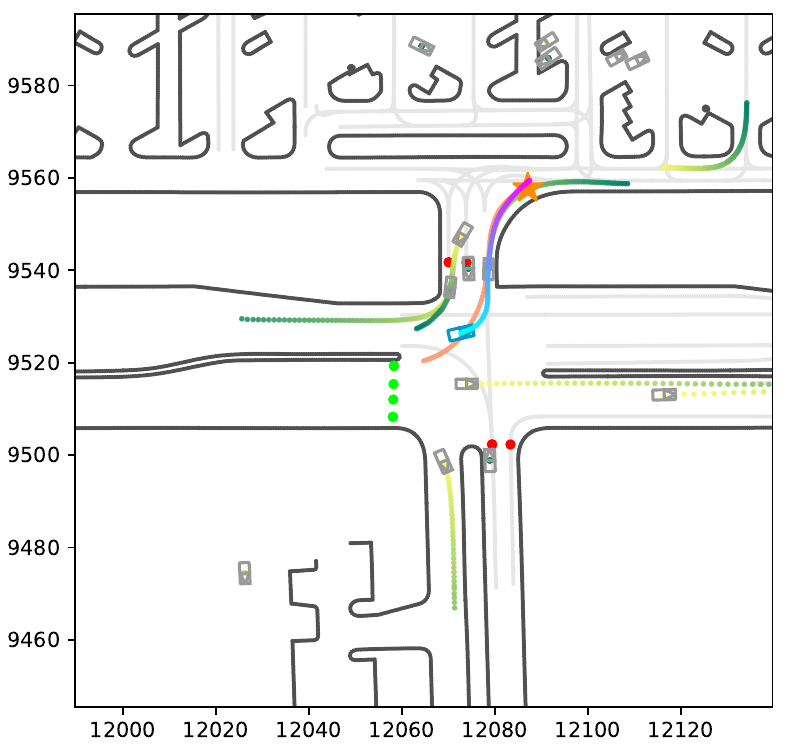}    \\
		                                                                                         & \addFig[trim=90 60 20 10, clip]{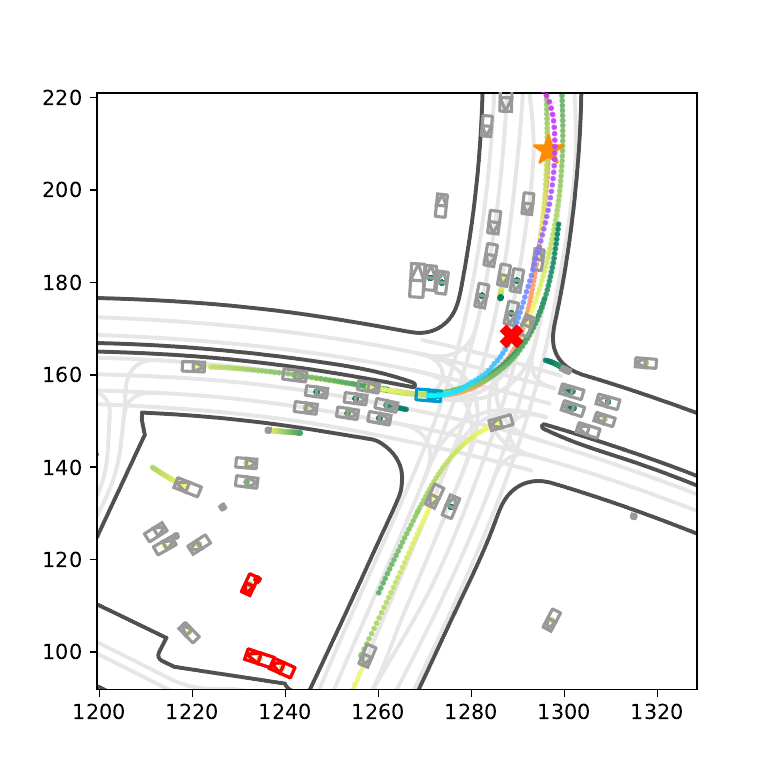}                        & \addFig[trim=90 60 20 10, clip]{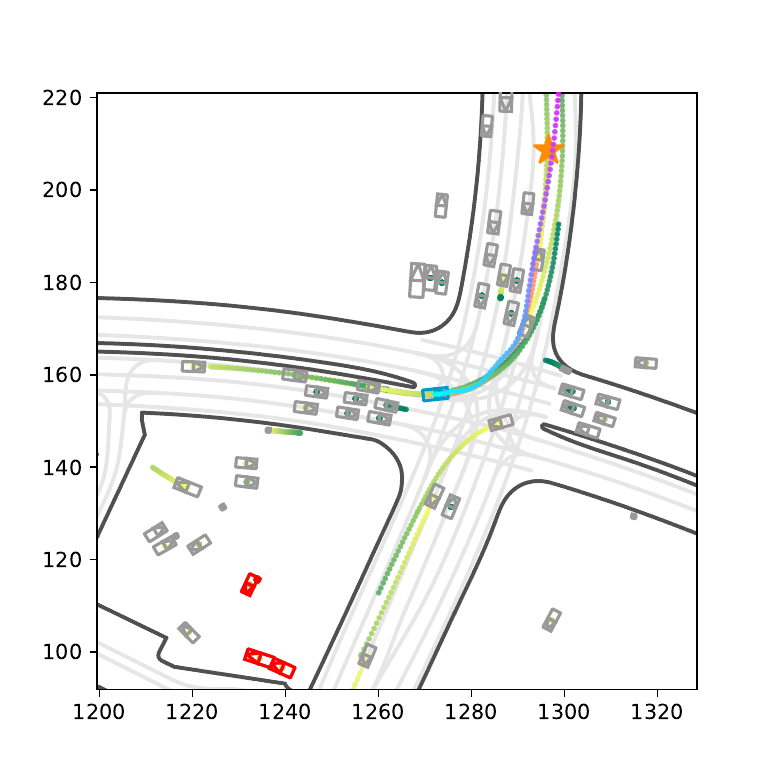}      & \addFig[trim=90 60 20 10, clip]{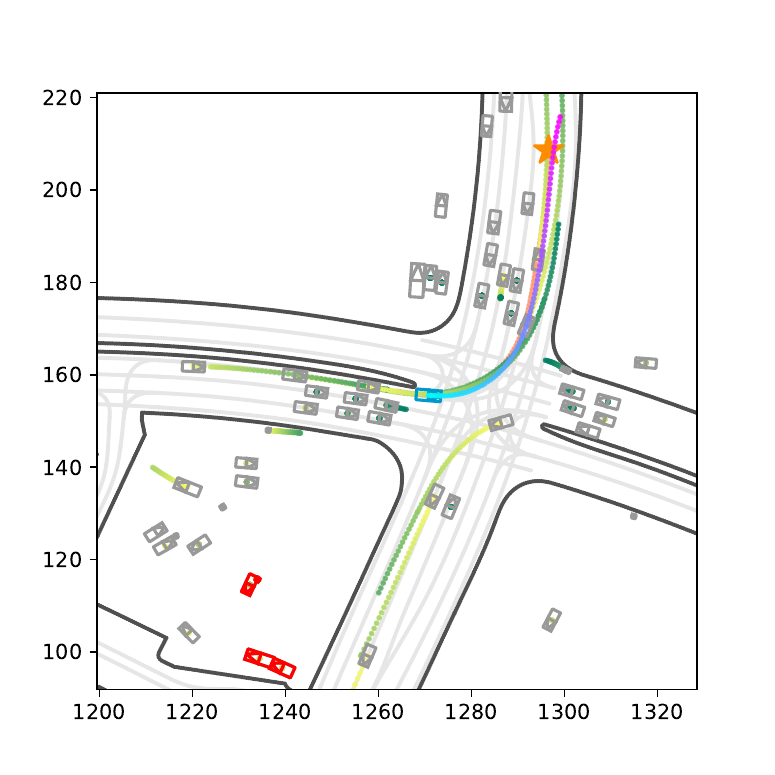}      & \addFig[trim=90 60 20 10, clip]{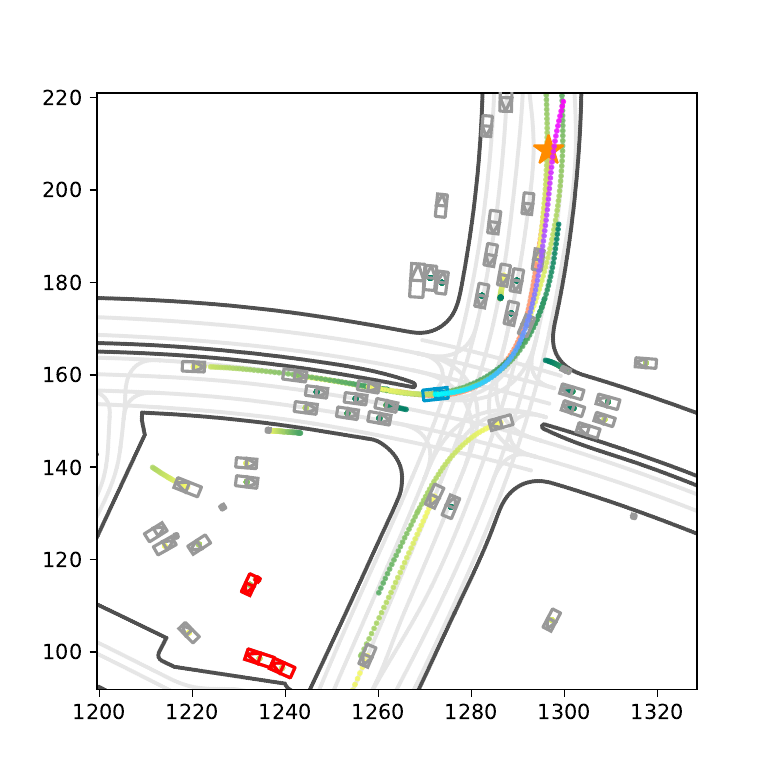}      \\
		                                                                                         & \multicolumn{4}{c}{\includegraphics[width=\linewidth]{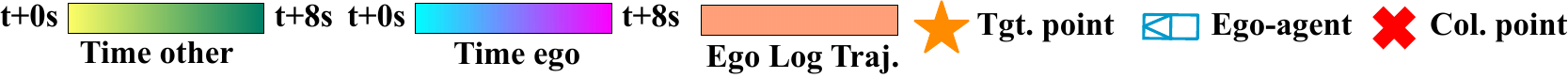}}
	\end{tabular}
	}
	\caption{Visualisation of \textit{close-loop} evaluation in representative urban
	scenarios is provided. For straight routing, demonstrations include controlled
	intersections, specifically, signalised intersections and roundabouts,
	presented in the first and second rows, respectively. In the scenario of the
	former, the ego-agent navigates straight into a complex intersection.
	Conversely, the latter scenario depicts the ego-agent following a leading vehicle
	into a congested roundabout. The third row features a T-junction where the ego-agent
	is required to execute multiple turns to enter a residential community. The concluding
	row displays the ego-agent executing a turn at a complex intersection while other
	vehicles are positioned in an adjacent lane, awaiting traffic light. }
	\label{fig.supp-urban}
\end{figure*}

We illustrate some representative \textit{close-loop} evaluation visualisations across
different driving scenarios in Figs.~\ref{fig.supp-lot}, \ref{fig.supp-highway-a}, \ref{fig.supp-highway-b},
and \ref{fig.supp-urban}. Vertically, we divide each driving scenario into two
parts based on the curvature of routing: straight and curvy. Horizontally, we compare
the performances of EasyChauffeur-IL and EasyChauffeur-PPO. Additionally, we provide
visualisations with Ego-Shifting, referring to `w/ ego-shifting', for
comprehensive comparison on each scenario. Fig.~\ref{fig.supp-lot} delves into
parking lot scenario, detailing the navigation through congestion, pedestrian
yielding, exiting into traffic, and executing minor-angle turns in tight spaces.
Fig.~\ref{fig.supp-highway-a} and \ref{fig.supp-highway-b} focus on dual carriageways, presenting high-speed travel
on straight roads amidst traffic, followed by navigating curvaceous and narrow
motorway sections, and exiting via slip roads. Fig.~\ref{fig.supp-urban} explores
urban driving, depicting straight routes into controlled intersections and
roundabouts, manoeuvring a T-junction into a residential area, and turning at
complex intersections with adjacent waiting vehicles. It can be seen that IL-trained
planner performs well on easy scenarios and fails to tackle the complex ones,
while the RL-trained planner presenting superior capability on various driving
settings and scenarios.

\newpage

